\documentclass[11pt]{article}

\usepackage[preprint]{acl}

\usepackage{times}
\usepackage{latexsym}
\usepackage{array}
\usepackage{booktabs}
\usepackage{enumitem}
\usepackage{multirow}
\usepackage{url}
 \usepackage{amsmath}
\usepackage[T1]{fontenc}
\usepackage[utf8]{inputenc}

\usepackage{microtype}

\usepackage{inconsolata}

\usepackage{graphicx}
\usepackage[utf8]{inputenc}
\usepackage{tcolorbox}
\tcbuselibrary{skins, breakable, listings}

\newtcblisting[use counter=figure]{promptbox}[2][]{%
    enhanced,
    float=h,           % Standard float placement
    width=\columnwidth, 
    colback=white,
    colframe=black,
    boxrule=0.5pt,
    arc=0mm,
    title={\textbf{Prompt:} #2}, 
    label={#1},
    listing only,
    listing options={
        basicstyle=\ttfamily\tiny, 
        breaklines=true,
        breakatwhitespace=true,
        columns=fullflexible,
        aboveskip=0pt,
        belowskip=0pt,
        extendedchars=true,  % Allow high-bit characters
        literate={’}{{'}}1 
             {‘}{{'}}1 
             {”}{{''}}1 
             {“}{{``}}1 
             {—}{{---}}1 
             {–}{{--}}1 
             {•}{{$\bullet$}}1 
             {…}{{...}}1
    }
}

\newtcblisting[use counter=figure]{promptboxwide}[2][]{%
    enhanced,
    float*=t,          % float* spans columns (usually needs 't' or 'b')
    width=\textwidth,  
    colback=white,
    colframe=black,
    boxrule=0.5pt,
    arc=0mm,
    title={\textbf{Prompt:} #2}, 
    label={#1},
    listing only,
    listing options={
        basicstyle=\ttfamily\tiny,
        breaklines=true,
        breakatwhitespace=true,
        columns=fullflexible,
        extendedchars=true,  % Allow high-bit characters
        literate={’}{{'}}1 
             {‘}{{'}}1 
             {”}{{''}}1 
             {“}{{``}}1 
             {—}{{---}}1 
             {–}{{--}}1 
             {•}{{$\bullet$}}1 
             {…}{{...}}1
    }
}

\NewDocumentCommand\emojione{}{\raisebox{-0.45em}{\includegraphics[height=2em]{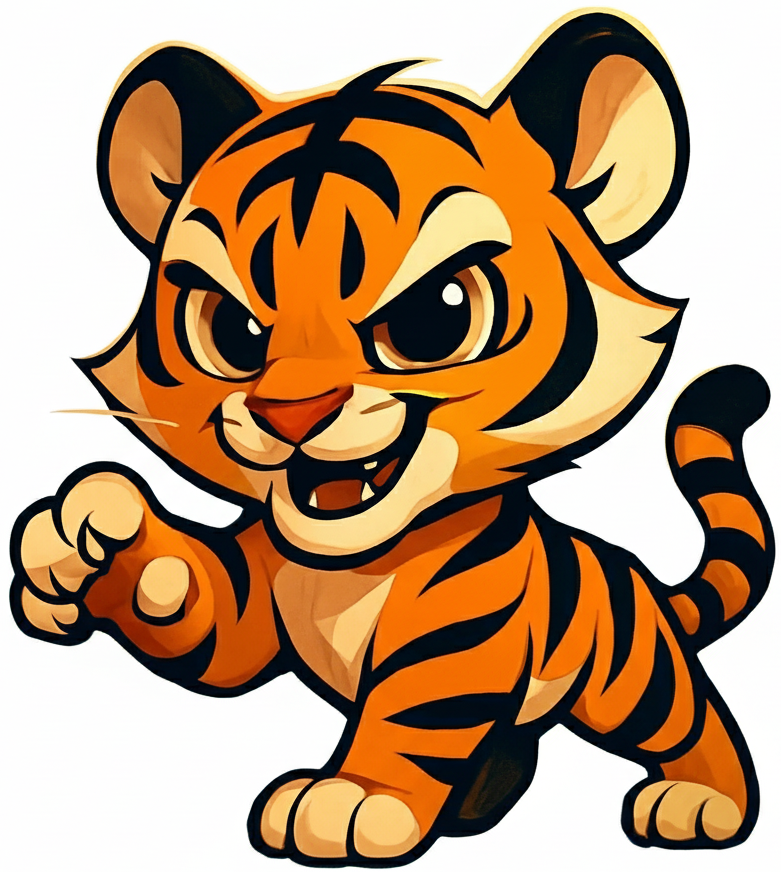}}}

\definecolor{bluehighlight}{RGB}{33,113,181}

\title{\emojione\hspace{0.01em} TAIGR: Towards Modeling Influencer Content on Social Media via Structured, Pragmatic Inference}
\author{
Nishanth Sridhar Nakshatri \quad
Eylon Caplan \quad
Rajkumar Pujari \quad
Dan Goldwasser \quad
    \vspace{0.1in} \\
    Department of Computer Science \\
    Purdue University \\
    {\tt \{nnakshat, ecaplan, rpujari, dgoldwas\}@purdue.edu}
}

\begin{document}
\maketitle
\begin{abstract}
Health influencers play a growing role in shaping public beliefs\footnote{\scriptsize \url{https://hsph.harvard.edu/news/maha-shaping-public-health-messages-on-social-media/}}, yet their content is often conveyed through conversational narratives and rhetorical strategies rather than explicit factual claims. As a result, claim-centric verification methods struggle to capture the pragmatic meaning of influencer discourse. In this paper, we propose \textbf{TAIGR} (\textbf{T}akeaway \textbf{A}rgumentation \textbf{I}nference with \textbf{G}rounded \textbf{R}eferences), a structured framework designed to analyze influencer discourse, which operates in $3$ stages: (1) identifying the core influencer recommendation$-$\textit{takeaway}; (2) constructing an argumentation graph that captures influencer justification for the \textit{takeaway}; (3) performing factor graph-based probabilistic inference to validate the \textit{takeaway}. We evaluate \textbf{TAIGR} on a content validation task over influencer video transcripts on health, showing that accurate validation requires modeling the discourse’s pragmatic and argumentative structure rather than treating transcripts as flat collections of claims.

\end{abstract}

\section{Introduction}

\begin{figure}[t!]
\centering
% width=0.48\textwidth, height=8cm
\includegraphics[scale=0.65]{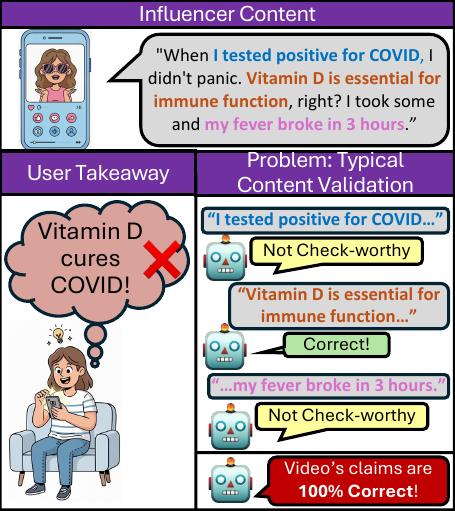}
  \small \caption{\small \textcolor{black}{Treating influencer content as a series of claims fails to account for the pragmatic \textit{Takeaway} conveyed to the user, and may lead to factually misleading results. We contend that un-``check-worthy'' \textit{anecdotes} may still be valuable to model, especially in their role of supporting an influencer's implicit takeaway.}}
  \vspace{-0.5cm}
  % Placeholder caption. Make it clear that this is the PROBLEM, not our proposed solution.}
  \label{fig:problem-explanation-figure}
\end{figure}

Social media influencers increasingly act as informal knowledge brokers~\cite{fabris2025self}, shaping public understanding and behavior in varied domains including health, finance, among others. Unlike traditional information sources, influencer-generated content is rarely presented as just factual reporting; instead, it is embedded in personal narratives, demonstrations and persuasive storytelling~\cite{lee_superstar_2021, georgakopoulou2024influencer, lim2025commercialization, evans2026unqualified}, often blurring the boundary between advice, belief and evidence. For instance, as shown in Figure~\ref{fig:problem-explanation-figure}, an objective claim like (``Vitamin D cures COVID'') is often woven into a personal story (``When I tested positive...''). Prior work has shown that such discourse can exert substantial influence on user attitudes and behavior, even when the users possess prior domain knowledge or skepticism~\cite{cialdini2007influence, https://doi.org/10.1002/mar.22173, evans2026unqualified}. In the high-stakes health domain, this creates a need for developing robust systems that do not just assess surface-level correctness, but can also analyze how influencer messages are constructed and justified.

% discretized \eylon{the reader may not know what you mean by discretized here. You mean that these are treated separate problems, independent of one another...} approach 
% as they are not typically organized around explicitly stated factual claims. 
From a linguistic perspective, influencer content exposes a gap in the current NLP approaches. Existing literature has approached related problems through claim verification~\cite{alhindi-etal-2018-evidence, wadden-etal-2020-fact, kotonya-toni-2020-explainable-automated, li-etal-2025-loki, dmonte_claim_2025}, argumentation mining~\cite{mochales2011argumentation, lawrence_argument_2019, hidey-etal-2017-analyzing} or social media modeling~\cite{bozdag2025readsystematicsurveycomputational, wang-etal-2019-persuasion, lee_superstar_2021}. However, treating these independently fails to provide a comprehensive view needed to understand the phenomenon of influencer discourse. An influencer's recommendation often manifests from anecdotes, credibility cues (``my doctor said'') or emotional framing; creating a gap between \textit{what an influencer says and what audience is encouraged to believe or do}. For example, in Figure~\ref{fig:problem-explanation-figure}, the influencer never explicitly says ``Vitamin D cures COVID'', yet the user is encouraged to believe it. Our experiments reveal that this gap is where claim-centric verification pipelines break down, as exemplified in Figure~\ref{fig:problem-explanation-figure}.

Rather than treating discourse as a flat collection of verifiable propositions, we model it closer to how humans interpret and evaluate communicated information. The theory of epistemic vigilance~\cite{sperber2010epistemic} posits that humans actively infer and evaluate communicated information through cognitive mechanisms that they use to protect themselves from being manipulated. These mechanisms operate along multiple dimensions: listeners infer \textit{what is being communicated}, evaluate \textit{the justification for the message}, and \textit{assess the support reliability}.
% , including source's credibility and consistency with prior knowledge. 
% \eylon{in this paragraph, bold the important parts that tie into the framework}

Inspired by this perspective, we propose a structured framework, \textbf{TAIGR}: \textbf{T}akeaway \textbf{A}rgumentation \textbf{I}nference with \textbf{G}rounded \textbf{R}eferences, to model influencer discourse. First, \textbf{TAIGR} identifies the key \textit{takeaway}, capturing \textbf{what} audiences internalize and act upon after watching influencer content. Then, it builds an argumentation structure, capturing \textbf{how} the influencer discourse supports the takeaway. Finally, augmenting this argumentation structure with external scientific evidence, it performs factor graph-based inference to \textbf{reason} about takeaway's trustworthiness.
To evaluate the efficacy of our approach in modeling influencer discourse, we adopt \textit{content validation} as the downstream evaluation task. This task serves as a holistic signal, requiring models to fully understand and reason about an influencer’s message. For this evaluation, we construct a dataset of $195$ real-world TikTok videos, grounded in expert annotations from \texttt{ScienceFeedback}\footnote{\scriptsize \url{https://science.feedback.org}}. In addition, we conduct targeted human evaluation to assess the quality of intermediate components of the framework. On this dataset, \textbf{TAIGR} outperforms all baselines, achieving gains of up to $+9.7$ macro-F1 points, highlighting the importance of structured modeling for analyzing influencer discourse.

% Additionally, utilizing the predictions from \textbf{TAIGR}, we analyze $1,430$ TikTok videos and uncover interesting patterns: (a) a video's medical accuracy is completely decoupled from virality. (b) video's trustworthiness is influenced by the rhetorical strategies employed by the influencer.
Beyond this evaluation, we apply \textbf{TAIGR} to $1{,}430$ medical-domain TikTok videos and uncover two key patterns: (i) trustworthiness is largely decoupled from platform popularity, and (ii) trustworthiness is strongly shaped by rhetorical strategy, with premise-based reasoning associated with higher trust and anecdotal narratives with lower trust. These findings illustrate how structured modeling enables influencer discourse analysis at scale. 

In summary, our contributions are:
\begin{itemize}[nosep]
    \item We propose \textbf{TAIGR}, a structured framework for analyzing influencer discourse, inspired by theories of epistemic vigilance.
    % content that decomposes discourse into \emph{what} is recommended, \emph{how} it is argued, and \emph{whether} it is supported by scientific evidence.
    \item We demonstrate that \textbf{TAIGR} significantly outperforms baselines on the \textit{content validation} task, achieving gains of up to $+9.7$ F1 points.
    \item We analyze influencer discourse on health at scale, showing that trustworthiness is independent of virality and is strongly influenced by rhetoric, revealing systematic differences between influencer and expert discourse.
\end{itemize}
\begin{figure*}[t!]
\centering
\hspace*{-0.5cm} 
% scale=0.605
\includegraphics[scale=0.51]{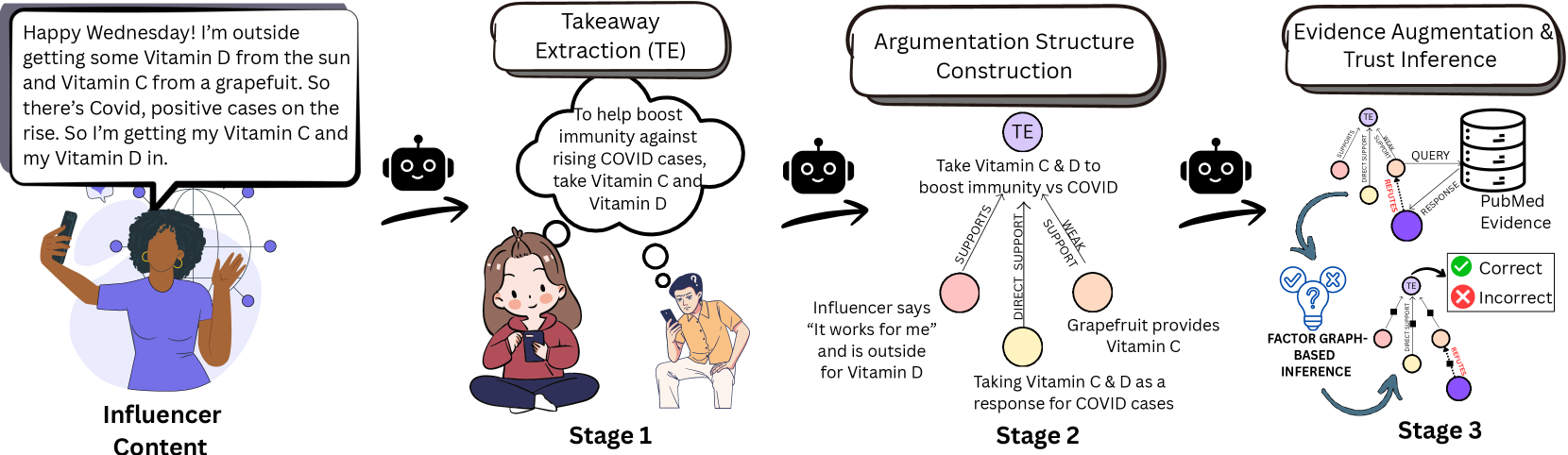}
  \scriptsize \caption{\small \textbf{TAIGR} overview: models influencer discourse in three stages: (1) \textit{Takeaway} extraction, identifying the core recommendation; (2) \textit{Argumentation Structure}, constructing a graph of supporting claims; and (3) \textit{Trust Inference}, validating the recommendation using external evidence.}
  \label{fig:main-framework}
  \vspace{-0.4cm}
\end{figure*}

\vspace{-0.1cm}
\section{Related Work}
\vspace{-0.1cm}
\label{sec:related-work}
Prior work relevant to modeling influencer-generated content has largely focused on isolated aspects of the problem, rather than modeling influencer discourse holistically.

\noindent \textbf{Claim Verification and Fact Checking.}
Early fact verification systems~\cite{vlachos-riedel-2014-fact, volkova-etal-2017-separating, rashkin-etal-2017-truth, perez-rosas-etal-2018-automatic}, primarily developed in the political domain, treat individual claims as the unit, without any external evidence. However, recent evidence-based methods either assume access to ``gold evidence''~\cite{alhindi-etal-2018-evidence, atanasova-etal-2020-generating-fact} or rely on generic retrieval pipelines~\cite{zheng-etal-2024-evidence, russo2025facefactsevaluatingragbased}. Similar approaches have been extended to scientific and health misinformation~\cite{wadden-etal-2020-fact, pradeep-etal-2021-scientific, kotonya-toni-2020-explainable-automated, vladika-etal-2024-healthfc, deka-etal-2023-multiple}. More recently, LLM-based systems have been adapted to do claim decomposition, retrieval, and verification~\cite{dmonte_claim_2025, li-etal-2025-loki, wang-etal-2025-openfactcheck}. However, these approaches remain largely \emph{claim-centric} and do not explicitly model the broader discourse structure through which influencers convey and justify health recommendations.
% central recommendation or the argumentative structure through which influencers justify it, both of which are crucial in modeling influencer discourse.

\noindent \textbf{Argumentation Mining and Reasoning Structures.}
A complementary line of work focuses on recovering the structure of arguments~\cite{mochales2011argumentation, lawrence_argument_2019, vecchi_towards_2021, sviridova_mining_2025}, and identifying claims and premises~\cite{boltuzic-snajder-2016-fill, hidey-etal-2017-analyzing, dusmanu-etal-2017-argument}. More recently, studies~\cite{zhu2025argragexplainableretrievalaugmented, ng2025margemeshingargumentativeevidence} leverage LLMs to construct and reason over argument graphs for claim verification. While these methods capture \emph{how} claims are supported, they are not developed to address the noisy, implicit, and persuasion-driven nature of influencer discourse.

\noindent \textbf{Social Media Modeling.}
Prior work examined how persuasion~\cite{wang-etal-2019-persuasion, lee_superstar_2021, akhynko-etal-2025-hidden, bozdag2025readsystematicsurveycomputational}, credibility~\cite{liu2024persuasive}, and influence~\cite{sharma_leveraging_2022, fang_using_2022} manifest in social media, offering valuable insights into rhetorical strategies.

In contrast to studies centered on social network analysis~\cite{del_tredici_you_2019, razis_modeling_2020}, our work focuses on the \emph{linguistic and argumentative structure} of influencer-generated content itself. By integrating takeaway identification, argumentation modeling, and evidence-grounded reasoning, we bridge insights from fact checking, argument mining, and social media analysis to analyze influencer discourse in an end-to-end manner.

\vspace{-0.1cm}
\section{TAIGR}
\vspace{-0.1cm}
Drawing inspiration from the theory of epistemic vigilance~\cite{sperber2010epistemic}, which characterizes how humans evaluate communicated information, we propose a structured framework, \textbf{TAIGR}, to analyze influencer discourse. It has three interdependent stages, examining different aspects of the discourse: (1) {Takeaway}: Influencer's key recommendation; (2) \textit{Argumentation Structure}: Represents the influencer's justification for the \textit{takeaway} using an argumentation graph; (3) \textit{Trust Inference}: Infers the degree to which the \textit{takeaway} is corroborated by external evidence;
% We present a structured framework for analyzing influencer discourse. The Theory of Epistemic Vigilance~\cite{sperber2010epistemic} provides a characterization of how humans evaluate communicated information. Building on this theory, we systematically examine three aspects of influencer content:
% \begin{enumerate}[topsep=0pt, itemsep=0pt, parsep=0pt]
%     \item \textit{Takeaway}: Influencer's key recommendation
%     \item \textit{Argumentation Structure}: Represents the influencer's justification for the \textit{takeaway} using an argumentation graph
%     % The set of claims and rhetorical strategies employed to support the Takeaway
%     \item \textit{Trust Inference}: Infers the degree to which the \textit{takeaway} is corroborated by external evidence
% \end{enumerate}

Figure~\ref{fig:main-framework} shows an overview of \textbf{TAIGR} and described in detail below. It operates over the text transcriptions of influencer videos, obtained using OpenAI's Whisper API\footnote{\scriptsize \url{https://pypi.org/project/openai-whisper/}}.

\subsection{Takeaway}
We define \textit{Takeaway} as the influencer's core actionable recommendation. It represents the central message the audience is intended to internalize and act upon, such as adopting a health routine, purchasing a product, or following a specific protocol.

% In other words, this takeaway constitutes the primary informational unit that audiences are encouraged to internalize and potentially act upon after watching the content. 

% expects the viewer to retain after consuming the content. 
The identification of \textit{takeaway} provides a semantic anchor for analyzing how the influencer supports or justifies the conveyed message. Similar to~\cite{alhindi-etal-2018-evidence}, who models the \emph{justification} of a claim to improve downstream performance, we argue that accurately identifying the \textit{takeaway} is a necessary prerequisite for effectively modeling influencer discourse.

% We conceptualize the \textit{takeaway} of an influencer video as the central recommendation or message that a viewer is likely to retain after consuming the content. It constitutes the primary informational unit that audiences are expected to internalize and potentially act upon after watching the content. In practice, they manifest from some actionable guidance, such as adopting a health routine, purchasing or avoiding a product, following a specific protocol. Therefore, we argue that accurately identifying them is critical for effectively modeling influencer content.

\begin{table}[t]
\centering
\scriptsize
\resizebox{\columnwidth}{!}{%
\begin{tabular}{p{6.5cm}|c}
\toprule
\multicolumn{1}{c|}{\textbf{Content}} & \textbf{Type} \\ 
\midrule
\textbf{Transcript-1:} [At urgent care] \textcolor{bluehighlight}{I was diagnosed with COVID}. [was sent home..called my nursing friends and doctor] \textcolor{bluehighlight}{They both recommended the ivermectin} [took $\text{1}^{\text{st}}$ dose..hoping] it relieves my symptoms soon. \par \medskip
% \textbf{Transcript-1:} Hey guys, day two of being in bed. Woke up yesterday morning with a high fever, body aches, congestion, even my teeth hurt. Went to CRMC urgent care where \textcolor{bluehighlight}{I was diagnosed with COVID}. \textit{[truncated for space]} I immediately started calling my nursing friends and my doctor. \textcolor{bluehighlight}{They both recommended the ivermectin}. I have been leery about taking it. \textit{[truncated for space]} I have went out on a limb and took my first dose today and hoping it relieves my symptoms soon. \par \medskip
\textbf{Takeaway:} If diagnosed with COVID, consider taking ivermectin as a potential treatment to relieve symptoms. & \multirow{10}{*}{Implicit} \\ 
\midrule
% \textbf{Transcript-2:} \textcolor{bluehighlight}{Go get your HPV vaccine}. Now, go get your daughter, your son, your niece, your nephew, your grandbaby, age 11 to 12. You can start as early as 9.  Get your HPV vaccine. \textit{[truncated for space]} This vaccine has been proven to prevent cancer. Okay. HPV related cancer, cervical cancer. Don't, don't let them take this opportunity away from you and your loved ones.  Go get it now. Okay. \par\medskip
\textbf{Transcript-2:} \textcolor{bluehighlight}{Go get your HPV vaccine}. Now, go get your [children], age 11 to 12. This vaccine has been proven to prevent [...] HPV related cancer, cervical cancer. \par \medskip
% \textcolor{bluehighlight}{Go get your HPV vaccine} [and get children] age 11 to 12. [It is] proven to prevent [...] HPV related cancer, cervical cancer. \par \medskip
\textbf{Takeaway:} Go get your HPV vaccine and ensure that your children aged 11 to 12 also receive it, as it has been proven to prevent HPV-related cancers such as cervical cancer. & \multirow{10}{*}{Explicit} \\ 
\bottomrule
\end{tabular}%
}
\caption{\small{Example for \textit{Implicit} vs. \textit{Explicit} takeaway type.}}
\label{tab:takeaway_sample_with_type}
\end{table}

Takeaways may be expressed either \textit{explicitly} or \textit{implicitly} (Table~\ref{tab:takeaway_sample_with_type}). Explicit takeaways are directly stated as advice or recommendations (e.g., `Get your vaccine'). Implicit takeaways, by contrast, are inferred from demonstrations, personal experiences, or causal narratives. In such cases, there is a gap between what the influencer says and what they imply. Despite their indirect form, they can be equally influential and persuasive.

Our takeaway extraction component captures both cases. Given the transcript of a video, we prompt an LLM to identify the key \emph{takeaway} from the transcript and classify whether it is \textit{explicit} or \textit{implicit}. Due to the short nature of the videos, we assume that each video contains exactly one \emph{takeaway}. Prompts used for Takeaway identification and classification are in Appendix~\ref{app:subsec-prompts-pipeline}.

% \subsection{\textit{Argumentation Structure}: How does the influencer support the \textit{takeaway}?}
\subsection{Argumentation Structure}
\label{subsec:argumentation-structure}
% Having identified the \textit{takeaway}, we develop a representation to analyze how the influencer argues in support of it. Rather than treating influencer content as a flat sequence of claims, we aim to recover the underlying argumentative structure that connects statements to the central takeaway. This structure captures not only \emph{what} is being claimed, but \emph{how} the influencer justifies the recommendation through supporting reasons and rhetorical strategies.
Having identified the \textit{takeaway}, we develop a representation to analyze how it is supported within the transcript by recovering its argumentative structure, linking claims to the central recommendation.

Following prior work in argumentation mining and claim--premise modeling~\cite{mochales2011argumentation, hidey-etal-2017-analyzing, sviridova_mining_2025}, we represent each transcript as a directed argumentation graph. We segment the transcript into standalone statements, assign rhetorical roles, identify claims in the transcript, and build the structure by inferring the relationship between them.

\begin{figure}[t!]
\includegraphics[width=\columnwidth]{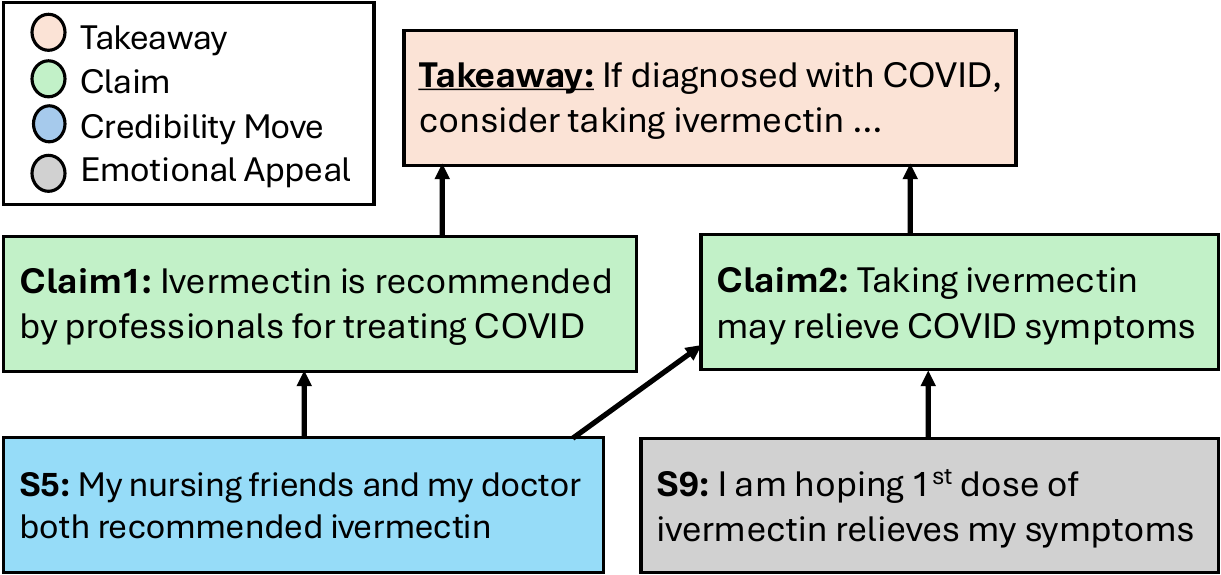}
  \small \caption{\small Argumentation graph for \textit{transcript-1} (Table~\ref{tab:takeaway_sample_with_type}), with node types and directed support edges.}
  % \eylon{typo in figure 1 ``nursing''. Also is there a better example of Emotional Appeal?}}
  \label{fig:arg-graph-example}
\end{figure}

\noindent \textbf{Notation.}
Let a transcript be denoted by a set of statements $\mathcal{S} = \{s_1, \dots, s_n\}$ and a set of claims $\mathcal{C} = \{c_1, \dots, c_m\}$. The extracted takeaway is denoted as $T$. Our objective is to construct a directed graph $G = (V, E)$, where $V = \mathcal{S} \cup \mathcal{C} \cup \{T\}$, and edges represent supportive argumentative relations.

\noindent \textbf{Preprocessing.}
Given a transcript, we first segment it into self-contained \emph{statements} by resolving co-references, using an LLM. These statements serve as the atomic units of influencer discourse.

Each statement $s_i \in \mathcal{S}$ is then classified by its rhetorical function, reflecting common persuasive strategies used by influencers. Following prior work~\cite{hidey-etal-2017-analyzing}, we label statements as one of: \textit{premise} (logos; appeals to logical reasoning), \textit{anecdotal evidence}, \textit{credibility move} (ethos), \textit{emotional appeal} (pathos), or \textit{None}. These labels are assigned using an LLM prompted to identify the argumentative role of each statement given the transcript. Figure~\ref{fig:arg-graph-example} shows an example of a statement $S5$ classified as \textit{credibility move}.

% Given a transcript, we first use an LLM to segment the text into standalone statements, resolving co-references so that each statement is self-contained. Each statement $s_i \in \mathcal{S}$ is then labeled according to its persuasive function, reflecting common rhetorical strategies in influencer content.

% Specifically, similar to~\cite{hidey-etal-2017-analyzing}, we classify statements into one of the following categories: \textit{premise} (logos$-$ appeals to reason), \textit{anecdotal evidence}, \textit{credibility move} (ethos), \textit{emotional appeal} (pathos), or \textit{None}. These labels are assigned using an LLM prompted to identify the argumentative role of each statement. 
% For example, in \textit{transcript-1} (Table~\ref{tab:takeaway_sample_with_type}), statement $S5$ is classified as a \textit{credibility move}, as shown in Figure~\ref{fig:arg-graph-example}.

To capture both explicit and implicit reasoning~\cite{sviridova_mining_2025}, we further extract a set of \emph{claims} $\mathcal{C}$, checkworthy propositions in the transcript, using an LLM. For each claim $c_j$, the LLM also identifies the subset of statements $\mathcal{S}_{c_j} \subseteq \mathcal{S}$ that support or give rise to the claim, yielding a local argumentative structure in which claims are justified by rhetorically typed statements. For instance, $Claim1$ is directly derived from $S5$ in Figure~\ref{fig:arg-graph-example}.
% This allows us to recover reasoning chains that are often implicit or enthymematic in influencer discourse.
% Next,  to capture reasoning chains that may be implicit or enthymematic~\cite{sviridova_mining_2025}, we identify a set of claims $\mathcal{C}$ made in the transcript, which are potentially checkworthy propositions made by the influencer. For each claim $c_j$, the LLM also identifies the subset of statements $\mathcal{S}_{c_j} \subseteq \mathcal{S}$ that are used to support or derive the claim, along with their rhetorical labels.

% This yields a local argumentative structure in which claim nodes are supported by one or more statement nodes. In Figure~\ref{fig:arg-graph-example}, for instance, $Claim1$ is directly derived from statement $S5$.

\noindent \textbf{Building the argumentation graph.}
We represent the influencer’s reasoning as a directed argumentation graph rooted at \textit{takeaway} node $T$. Nodes correspond to the extracted argumentative units (claims, statements, takeaway), and directed edges represent inferred support relations between them.

We begin by identifying candidate edges from each claim $c_j \in \mathcal{C}$ to the takeaway $T$. For each pair $(c_j, T)$, we prompt an LLM to classify the relationship as \textit{direct support}, \textit{weak support}, or \textit{no support}. Following prior work showing LLMs can meaningfully express uncertainty~\cite{kadavath2022language}, we compute a continuous edge weight using the model’s class probabilities: $w(c_j, T) = 1.0 * P(\text{direct support}) + 0.5 * P(\text{weak support})$. Edges whose scores exceed a fixed threshold are added to the graph, yielding an initial set of claim-to-takeaway support links.

We then iteratively expand the graph to capture deeper argumentative structure. For each node already connected to the graph, we evaluate whether remaining claims provide support to it, using the same classification and weighting procedure. Edges that meet the threshold are added, allowing claims to support intermediate claims as well as the takeaway.

Finally, we attach remaining statement-level nodes using the same process, continuing expansion until all candidate nodes have been evaluated.

The resulting graph captures the inferred structure of how an influencer’s claims collectively support the takeaway. Figure~\ref{fig:arg-graph-example} illustrates an example argumentation graph for \textit{transcript-1} (Table~\ref{tab:takeaway_sample_with_type}). 

Implementation details, thresholds, and prompts are provided in Appendix~\ref{app:subsec-hyperparameters-implmentation-details} and Appendix~\ref{app:subsec-prompts-pipeline}.

\subsection{Trust Inference}
\vspace{-0.1cm}
After constructing the argumentation graph, we estimate whether the \textit{takeaway} should be trusted by assessing the factual accuracy of the arguments used to support it. Rather than modeling trust as an intrinsic property of the influencer, we operationalize it as the degree to which the supporting arguments align with external scientific evidence. Under this formulation, a takeaway is considered more trustworthy when its argumentative support is consistent with reliable biomedical literature.

We validate using two steps. First, we retrieve and classify scientific evidence from~PubMed\footnote{\scriptsize \url{https://pubmed.ncbi.nlm.nih.gov/}} to augment the argumentation graph. Second, we reason over this evidence-augmented graph to infer the trustowrthiness of the \textit{takeaway}.

\subsubsection{Evidence Retrieval and Classification}
Our evidence retrieval pipeline follows~\cite{li-etal-2025-loki} and consists of four steps: identifying checkworthy nodes, expanding search queries, retrieving and reranking evidence, and classifying evidence; described below in detail. Implementation details and prompts are provided in Appendix~\ref{app:subsec-hyperparameters-implmentation-details} and~\ref{app:subsec-prompts-pipeline}.
% Similar to the recent approach for claim verification~\cite{li-etal-2025-loki}, our evidence retrieval pipeline involves: (1) identifying checkworthy nodes in the argumentation graph; (2) expanding the search queries to improve recall; (3) retrieving and reranking evidence; and (4) classifying evidence. Implementation specific details on PubMed data indexing and relevant prompts are provided in Appendix~\ref{app:subsec-hyperparameters-implmentation-details}, ~\ref{app:subsec-prompts-pipeline}, respectively.

\noindent \textbf{Checkworthy Node Identification.}
Not all nodes in the argumentation graph are verifiable. We restrict evidence retrieval to \textit{Claim} and \textit{Premise} nodes, excluding rhetorical categories such as anecdotal evidence or emotional appeals. For each candidate node, an LLM determines whether it is \emph{PubMed-checkworthy}, i.e., whether it expresses a proposition that can plausibly be supported or refuted by biomedical literature.
% \textbf{Identifying Checkworthy Nodes.} Not every node in the argumentation graph can be checked with external verification. Especially, nodes corresponding to \textit{\{Anecdotal Evidence, Credibility Move, Emotional Appeal\}}. Therefore, we restrict external verification for \textit{Claim} and \textit{Premise} node types. For each such node, we prompt an LLM to determine whether it is ``PubMed-checkworthy'', i.e., whether it expresses a proposition that can plausibly be supported or refuted by biomedical literature.

% \textbf{Expanding Search Queries.} To improve recall and to surface both supporting and opposing evidence, we expand each checkworthy node into a set of search queries, following~\cite{ousidhoum-etal-2022-varifocal}. For every node, we prompt an LLM to generate up to five supporting queries and up to five opposing queries. This bidirectional expansion encourages retrieved evidence to reflect both scientific consensus as well as contradictory findings, when they exist.
\noindent \textbf{Query Expansion.}
To improve recall and capture both supporting and opposing evidence~\cite{ousidhoum-etal-2022-varifocal}, we expand each checkworthy node into multiple queries. Specifically, an LLM generates up to five supporting and five opposing queries per node, encouraging retrieval of evidence with diverse scientific stances.

% we prompt an LLM to obtain up to five supporting queries and up to five opposing queries. For instance, if the query is ``Topical aged urine increases testosterone and improves skin appearance.'', then ``Effects of topical application of human urine on skin hydration and color'' is a supporting query; and ``Clinical trials showing no testosterone increase after topical application of human urine'' is an opposing query.

% \textbf{Evidence Retrieval and Reranking.}
% For each query in the expanded set, we retrieve the top-$K$ ($K=100$) candidate documents from PubMed using \texttt{ColBERT-v2}~\cite{santhanam-etal-2022-colbertv2}, a dense retriever optimized for semantic search. While dense retrieval is effective at identifying topically related documents, at times, it also retrieves evidence that is only weakly relevant to the specific claim being evaluated. To address this limitation, we apply a \texttt{monoT5} reranker~\cite{nogueira-etal-2020-document} (\texttt{monot5-base-med-msmarco}) to perform fine-grained relevance matching between each claim and the retrieved evidence. After reranking, we retain the top-$30$ evidence documents across all queries for evidence classification.

\noindent \textbf{Evidence Retrieval and Reranking.}
For each expanded query, we retrieve the top-$K$ ($K{=}100$) candidate PubMed documents using \texttt{ColBERT-v2}~\cite{santhanam-etal-2022-colbertv2}. To filter weakly relevant results, we apply a \texttt{monoT5} reranker~\cite{nogueira-etal-2020-document} (\texttt{monot5-base-med-msmarco}) and retain top-$30$ evidence documents per node for downstream classification.

% \textbf{Evidence Retrieval and Reranking.} For each query from the expanded set (including the original query), we retrieve top-K ($K=100$) evidence using \texttt{ColBERT-v2}~\cite{santhanam-etal-2022-colbertv2}. We merge the retrieved evidence and retain only top-K based on the retrieved scores. While dense retrievers such as ColBERT-v2 offer high recall, they often retrieve topically related but weakly relevant documents. Therefore, we apply a \texttt{monot5-base-med-msmarco}\footnote{\url{https://huggingface.co/castorini/monot5-base-med-msmarco}}~\cite{nogueira-etal-2020-document} reranker to perform fine-grained relevance matching between claims and evidence. Overall, we retain only the top-30 evidence articles for a node.

\noindent \textbf{Evidence Classification.}
Each retrieved PubMed article is classified by an LLM into one of five categories:
\{\textit{strong support, weak support, neutral, weak oppose, strong oppose}\}.
Along with the class label, we additionally compute a continuous edge weight using the LLM’s class probabilities: $e = -1.0 \cdot P(\text{strong oppose}) - 0.5 \cdot P(\text{weak oppose}) + 0 \cdot P(\text{neutral}) + 0.5 \cdot P(\text{weak support}) + 1.0 \cdot P(\text{strong support})$. This formulation jointly captures the polarity and strength of the evidence, yielding a score in $[-1,1]$ that quantifies how strongly an evidence node supports or contradicts the associated claim or premise.

The resulting evidence nodes and weighted support or attack edges are added to the argumentation graph, producing an evidence-augmented structure, used for inferring the takeaway's trustworthiness.

\subsubsection{Factor graph based content validation}
Given the evidence-augmented argumentation graph, we infer how much the influencer’s \textit{takeaway} should be trusted by propagating evidence credibility through the argumentative structure. We treat PubMed evidence as the most reliable source and use it as an anchor from which trust propagates to claims and ultimately to the takeaway.

To enforce global consistency over this structure, we formulate trust inference as a factor graph and perform approximate inference using AD3~\cite{martins2015ad3}. This formulation allows us to jointly reason about local evidence strength and global argumentative relations, while explicitly modeling both support and contradiction.

\noindent \textbf{Variables and states.}
Each node $v$ in the augmented argumentation graph corresponds to a discrete random variable $Y_v$, representing its trustworthiness. Trust is discretized into $11$ states $\{0, 0.1, \ldots, 1.0\}$. PubMed evidence nodes are assigned a hard constraint $Y_v = 1.0$, reflecting their assumed high credibility.

\noindent \textbf{Inference objective.}
We seek a trust assignment that maximizes the sum of unary and binary potentials:
\[
\arg\max_{y_1,\ldots,y_m}
\sum_{v \in V} \theta_v(y_v) + \sum_{(u,v) \in E} \theta_{uv}(y_u, y_v),
\]
where unary potentials encode prior trust assumptions and binary potentials specify how trust values of connected nodes should influence one another.

\noindent \textbf{Unary potentials.}
Unary potentials enforce hard constraints for PubMed evidence nodes, and encode lower prior trust for rhetorical nodes via soft potentials. All remaining nodes receive uniform unary potentials, allowing their trust to be determined by surrounding evidence and structure.
% Unary potentials enforce hard constraints for PubMed evidence nodes and assign lower prior trust to rhetorical or non-factual nodes. All remaining nodes receive uniform unary potentials, allowing their trust to be determined by surrounding evidence and structure.

\noindent \textbf{Binary potentials.}
Binary potentials define how trust flows between connected nodes in the argumentation graph, with stronger support or opposition relations leading to greater influence on the final trust assignment.

For a \textbf{support} edge, we define:
$\theta^{\text{support}}(y_u, y_v) = \beta \cdot w \cdot \bigl(1 - (y_u - y_v)^2\bigr)$,
which encourages similar trust assignments between supporting nodes. As a result, trust from credible evidence propagates through supporting claims towards the takeaway.

For an \textbf{attack} edge, we define: $\theta^{\text{attack}}(y_u, y_v) = \gamma \cdot |w| \cdot \bigl(1 - (y_u + y_v - 1)^2\bigr)$,
which encourages complementary trust assignments. This formulation ensures strong trust in one node implies reduced trust in the opposing node, preventing contradictory evidence from jointly receiving high confidence.

Here, $\beta$ and $\gamma$ control the relative influence of support and attack relations\footnote{\scriptsize We set $\beta = 0.3, \gamma = 5$.}.

\noindent \textbf{Inference and validation.}
We solve the resulting factor graph using AD3, which relaxes the discrete problem into a linear program and efficiently produces approximate marginal trust assignments. The inferred trust value of the \textit{takeaway} node serves as our content validity score: values in $[0, 0.5]$ are labeled \textit{incorrect}, $[0.5, 0.7]$ as \textit{partially correct}, and $> 0.7$ as \textit{correct}. Implementation details and hyperparameters are provided in Appendix~\ref{app:subsec-hyperparameters-implmentation-details}.

\vspace{-0.1cm}
\section{Data}
\vspace{-0.2cm}
\label{sec:data-sec}
Following~\cite{shang_multitec_2025}, we use TikTok's Research API to construct a dataset of health-related videos via hashtag-based search. The hashtags are derived from \texttt{ScienceFeedback} website, ensuring we primarily capture health-related influencer discourse (see Appendix~\ref{app:subsec-hyperparameters-implementation-details-dataset} for details). To focus on influencer-generated content with meaningful audience reach, we retain only videos posted by creators with at least $1{,}000$ followers. The resulting dataset consists of $1{,}430$ videos. Table~\ref{tab:data-distribution} summarizes the distribution of videos by influencer popularity and by takeaway type. We note that a substantial portion of the dataset contains \textit{implicit} takeaways, reflecting the indirect nature of health advice commonly observed in influencer content.

\begin{table}[h]
\centering
\resizebox{\columnwidth}{!}{
\begin{tabular}{p{35mm} l @{\hspace{8mm}} p{35mm} l}
\toprule

\multicolumn{4}{l}{\textit{By Influencer Popularity}} \\ \hline

\textbf{Category} & \textbf{\#Videos} &
\textbf{Category} & \textbf{\#Videos} \\

\hspace{3mm} 0 -- 50K     & 603 &
\hspace{3mm} 250K -- 500K & 155 \\

\hspace{3mm} 50K -- 100K  & 218 &
\hspace{3mm} 500K -- 1M   & 112 \\

\hspace{3mm} 100K -- 250K & 220 &
\hspace{3mm} $>$ 1M       & 122 \\

\midrule
\multicolumn{4}{l}{\textit{By Takeaway Type}} \\
\midrule

\hspace{3mm} Implicit      & 533 & \multicolumn{2}{l}{} \\
\hspace{3mm} Explicit      & 897 & \multicolumn{2}{l}{} \\

\midrule
\textbf{Total Dataset}     & \textbf{1,430} & \multicolumn{2}{l}{} \\
\bottomrule
\end{tabular}
}
\caption{\small{Dataset Distribution.}}
\label{tab:data-distribution}
\end{table}

% \begin{table}[h]
% \small
% \centering
% \begin{tabular}{lc}
% \toprule
% \textbf{Influencer Popularity} & \textbf{Video Count} \\
% \midrule
% 0 to 50K       & 603   \\
% 50K to 100K    & 218   \\
% 100K to 250K   & 220   \\
% 250K to 500K   & 155   \\
% 500K to 1M     & 112   \\
% $>$ 1M         & 122   \\
% \midrule
% \textbf{Implicit} takeaway  & 533 \\
% \textbf{Explicit} takeaway & 897 \\
% \textbf{Total} & \textbf{1,430} \\
% \bottomrule
% \end{tabular}
% \caption{\small{Dataset distribution of videos based on influencer popularity.}}
% \label{tab:data-distribution}
% \end{table}
\vspace{-0.1cm}
\section{Evaluation}
\vspace{-0.1cm}
\begin{table*}[t!]
\centering
% \scriptsize
\resizebox{0.8\linewidth}{!}{%
\begin{tabular}{>{\arraybackslash}m{5cm}|>{\centering\arraybackslash}m{2.2cm}|>{\centering\arraybackslash}m{1.6cm}>{\centering\arraybackslash}m{1.6cm}>{\centering\arraybackslash}m{1.6cm}|>{\centering\arraybackslash}m{1.6cm}>{\centering\arraybackslash}m{1.6cm}}
\toprule
\multirow{2}{*}{\textbf{Method}} & 
\multirow{2}{2.2cm}{\centering \textbf{F1} \\ \textbf{(Macro)}} & 
\multicolumn{3}{c|}{\textbf{Class-wise F1}} & 
\multicolumn{2}{c}{\textbf{F1 by Takeaway Type}} \\
\cmidrule(lr){3-5} \cmidrule(l){6-7}
 & & \textbf{Incorrect} & \textbf{Partial} & \textbf{Correct} & \textbf{Implicit} & \textbf{Explicit} \\ 
\midrule
\textbf{Claim-centric RAG} & $0.3$ & $0.05$ & $0.32$ & $0.53$ & $0.21$ & $0.31$ \\
\textbf{RAG-w-Takeaway} & $0.42$ & $0.2$ & $0.39$ & $0.69$ & $0.44$ & $0.41$ \\
\textbf{LOKI}~\cite{li-etal-2025-loki} & $0.43$ & $0.71$ & $0.29$ & $0.29$ & $0.35$ & $0.43$ \\
\midrule
\textbf{TAIGR (No Structural Propagation)} & $0.46$ & $0.66$ & $0.23$ & $0.47$ & $0.47$ & $0.43$ \\
\textbf{TAIGR} & $\textbf{0.52}$ & $0.72$ & $0.42$ & $0.41$ & $\textbf{0.65}$ & $\textbf{0.49}$ \\
\bottomrule
\end{tabular}}
\caption{\small{Content validation task results. \textbf{(a) Main Results:} The first columns show the overall Macro-F1 and the class-wise performance breakdown (Incorrect/Partial/Correct). We assessed statistical significance using a paired bootstrap test ($10000$ iterations). Testing the hypothesis that our model outperforms \texttt{LOKI}, we achieved a statistically significant result ($p=0.039 < 0.05$). \textbf{(b) Takeaway Analysis:} The rightmost columns compare performance across takeaway type.}}
\label{tab:combined_results}
\end{table*}
We evaluate our framework using human annotations and an automated \textit{content validation} task, which requires a model to judge whether an influencer’s content is \{\textit{correct, partially correct, incorrect}\} given the video transcript.

Crucially, solving this task requires more than surface-level pattern matching: a model must recover the influencer’s intended recommendation, understand how it is argued for, and assess whether the supporting claims are scientifically valid. We therefore use content validation task as an end-to-end evaluation signal for the proposed framework. We use \texttt{GPT-4.1-mini}\footnote{\scriptsize\url{https://platform.openai.com/docs/models/gpt-4.1-mini}} for all LLM calls and report results from a single execution of the model.

% We note that we use \texttt{GPT-4.1-mini}\footnote{\scriptsize\url{https://platform.openai.com/docs/models/gpt-4.1-mini}} as the model for all LLM calls, and report results based on a single LLM execution.

\subsection{Experimental Setup}
\paragraph{Annotated Dataset.}
To support content validation, we construct an annotated dataset by aligning influencer videos with expert-reviewed medical claims. We use \texttt{ScienceFeedback}, which contains real-world claims verified by medical experts and labeled into $18$ fine-grained veracity categories.

For each video transcript (Section~\ref{sec:data-sec}), we extract its central \textit{takeaway} and identify semantically corresponding claims from \texttt{ScienceFeedback}. We retain pairs where the expert-annotated claim entails the extracted takeaway, yielding $185$ candidates. After manual verification to ensure alignment semantic equivalence, we obtain $166$ high-quality \{\textit{takeaway}, claim\} pairs.

Due to limited data and label skewness, we collapse the $18$ original categories into three coarse-grained classes: \{\textit{correct, partially correct, incorrect}\}, and assign each video the label of its associated takeaway. To increase coverage of high-confidence correct cases, we further augment the dataset with $29$ videos from the \texttt{Cleveland Clinic}, a medically authoritative source, all labeled as \textit{correct}. In total, the evaluation set contains $195$ labeled video transcripts. Additional details are provided in Appendix~\ref{app:subsec-constructing-annotated-dataset}.
\paragraph{Baselines.}
We compare our framework against the following baselines (detailed in Appendix~\ref{app:subsec-baselines}), which differ in how explicitly they model influencer content and its underlying reasoning.

\textbf{Claim-centric RAG.}
This baseline decomposes a transcript into factual claims using an LLM, retrieves PubMed evidence for each claim, and classifies the retrieved evidence as supporting or opposing. The LLM then predicts the overall veracity of the transcript based on the aggregated evidence. This setup does not explicitly model the central recommendation conveyed by the influencer.

\textbf{RAG-w-Takeaway.}
This baseline introduces limited structure by first extracting the \textit{takeaway} and retrieving PubMed evidence for it. The retrieved evidence is classified as supporting or opposing, and the LLM is prompted to predict the veracity of the \textit{takeaway} given this evidence. This baseline isolates the impact of takeaway-aware retrieval without modeling claim interactions.
% isolates the effect of takeaway-aware retrieval while omitting explicit modeling of claims and their interactions.

\textbf{LOKI.}
We adapt LOKI~\cite{li-etal-2025-loki}, a claim-centric verification system, to our setting. LOKI decomposes transcripts into claims, filters uncheckworthy ones, expands retrieval queries to improve recall, and verifies each claim independently using PubMed evidence. We follow~\cite{li-etal-2025-loki} for threshold selection for the \textit{incorrect} class and tune the \textit{partially correct} threshold (Appendix~\ref{app:subsec-baselines}).

\paragraph{Metrics.} We evaluate all models using a standard classification metric: macro-averaged F1-score, to account for class imbalance.

\subsection{Results and Ablation}
Table~\ref{tab:combined_results} presents results on the content validation task. Overall, our framework achieves the highest macro F1-score, outperforming all competing baselines by a margin of up to $+9.7$ points.

\paragraph{Main Results.}
Among the baselines, \textbf{Claim-centric RAG} performs poorly, indicating that treating influencer transcripts as collections of loosely related factual claims is insufficient for validating their overall message. Introducing minimal structure via \textbf{RAG-w-Takeaway} substantially improves performance, highlighting the importance of explicitly modeling what the influencer is recommending.

% In contrast, \textbf{RAG-w-Takeaway} substantially improves performance, demonstrating that explicitly modeling what the influencer is recommending provides a strong signal for downstream verification.

\textbf{LOKI} performs comparably to \textbf{RAG-w-Takeaway}, despite using a more sophisticated claim-filtering and query-generation pipeline. However, both baselines fall short of our method. While LOKI verifies claims individually and aggregates them uniformly, it does not model their interdependencies or relative influence on the final takeaway. \textbf{TAIGR} explicitly captures these dependencies, enabling more accurate trust inference. The improvement over the strongest baseline (LOKI) is statistically significant under a paired bootstrap test with $10{,}000$ iterations ($p = 0.039 < 0.05$).

% \textcolor{red}{Assumption-based: Compared to \textbf{Direct Prompting}, our method yields a substantial performance gain, highlighting the limitations of end-to-end prompting for complex influencer content. This result indicates that accurately validating influencer recommendations requires explicit modeling of intermediate structures, including the central takeaway, supporting arguments, and external evidence, rather than relying solely on implicit reasoning within a single prompt.}

% Among structured baselines, \textbf{RAG-w-Takeaway} performs comparably to \textbf{LOKI}, suggesting that explicitly modeling what the influencer is recommending is already a strong signal for content validation. However, both baselines fall short of our framework. While LOKI decomposes transcripts into claims, it treats claims independently and aggregates them uniformly, without modeling their interdependencies or relative influence on the final recommendation. In contrast, our method explicitly captures how claims support one another and the takeaway, resulting in more accurate trust inference.

% We assess statistical significance using a paired bootstrap test with $10{,}000$ iterations. The improvement of our method over the strongest baseline (LOKI) is statistically significant ($p = 0.039 < 0.05$).

\begin{figure}[t!]
\centering
\includegraphics[scale=0.42]{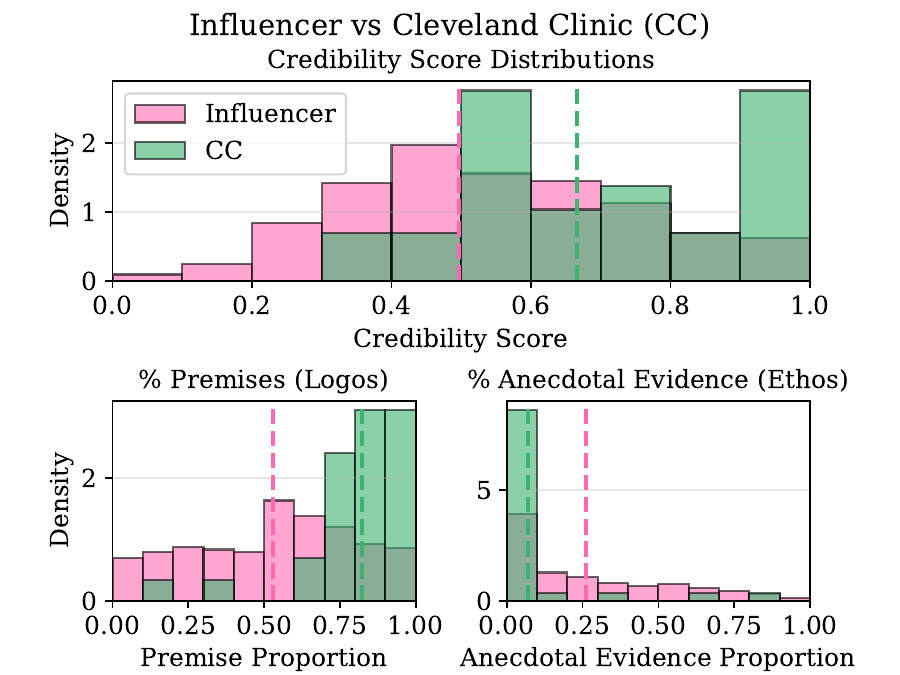}
  \small \caption{\small Rhetorical distributions of TikTok influencers vs. experts: expert content relies on premises and achieves higher credibility, while influencer content emphasizes anecdotes.}
  % \small \caption{\small What drives the credibility gap between influencers and experts? We compare rhetorical distributions of TikTok vs. the Cleveland Clinic: experts achieve higher credibility by relying on logical premises, whereas influencers rely heavily on anecdotal evidence.}
  \label{fig:tiktok_cc_comparison}
\end{figure}

\noindent \textbf{Impact of Takeaway Type.}
Table~\ref{tab:combined_results}(b) shows performance by takeaway type. Our method outperforms all baselines for both \textit{explicit} and \textit{implicit} takeaways, with particularly large gains for implicit cases. For implicit takeaways, the improvement over LOKI is statistically significant (Appendix~\ref{app:stat-significance}).

This gap is primarily attributable to LOKI’s reliance on surface-level claim decomposition. Influencer transcripts are conversational and often imply recommendations indirectly. When the takeaway is implicit, LOKI frequently fails to recover a claim that corresponds to the intended recommendation, leading to ineffective evidence retrieval. Table~\ref{tab:loki_failure_analysis} illustrates this failure mode: for \textit{transcript-1}, LOKI extracts claims centered on peripheral details (e.g., who recommended ivermectin), but fails to capture the implied recommendation of using ivermectin to treat COVID-19. Consequently, relevant evidence is not retrieved, resulting in incorrect validation.

\noindent \textbf{Impact of Argument Structure Modeling.}
We evaluate the effect of modeling interactions between claims by comparing \textit{\textbf{TAIGR} (No Structural Propagation)}, which performs independent claim validation without graph-based inference, against our full method. As shown in Table~\ref{tab:combined_results}(a), this variant outperforms LOKI, reflecting improved claim extraction and evidence retrieval. However, it remains substantially worse than the full model ($-6$ F1 points), demonstrating the importance of modeling claim dependencies and trust propagation through the argumentation graph.
% that modeling inter-claim dependencies and propagating trust through the argumentation graph is critical for accurate trust inference.
% To isolate the effect of modeling interactions between claims, we evaluate a variant of our framework that validates claims independently, without factor graph-based trust inference (\textit{TAIGR (No Structural Propagation)}). As shown in Table~\ref{tab:main-results}, this variant outperforms LOKI, reflecting improved claim extraction and evidence retrieval. However, it remains substantially worse than the full model ($-6$ F1 points), demonstrating that modeling inter-claim dependencies and propagating support through the argumentation structure is critical for accurate trust inference.

Taken together, these results show that performance gains arise not from any single component, but from the integration of (i) takeaway-based modeling, (ii) argumentation structure, and (iii) structured trust propagation over retrieved evidence.

% \noindent \textbf{Human Validation}
% In addition to the automated evaluation, on a randomly sampled $100$ examples, we conduct human validation as a qualitative sanity check, assessing the reliability of intermediate components in our framework. Specifically, we evaluate the accuracy of the LLM in identifying: (a) the correct \textit{takeaway}, (b) whether the takeaway is expressed \textit{implicitly} or \textit{explicitly}, and (c) the rhetorical labels of individual statements in the transcript (four classes). As shown in Figure~\ref{fig:human-eval-precision}, the extracted takeaways and rhetorical labels are largely consistent with human judgments. These results suggest that LLMs can recover meaningful high-level semantic and rhetorical structure from noisy, conversational influencer discourse.

\noindent \textbf{Human Validation.}
As a qualitative sanity check of our framework’s intermediate components, we conduct human validation on a randomly sampled set of $100$ videos. We evaluate the LLM’s ability to (i) correctly identify the video’s \textit{takeaway}, (ii) determine whether the takeaway is expressed \textit{explicitly} or \textit{implicitly}, and (iii) assign rhetorical function labels to individual statements. Human judgments show that the LLM identifies the takeaway with $75\%$ accuracy, classifies implicit versus explicit takeaways with $79\%$ accuracy, and assigns rhetorical labels with $72\%$ accuracy. Despite the noisy and conversational nature of influencer discourse, these results indicate that LLMs can reliably recover the semantic intent and rhetorical structure required by our framework.
\vspace{-0.1cm}
\section{Analysis of Rhetoric and Credibility}
\vspace{-0.2cm}
\begin{figure}[t!]
\centering
\includegraphics[scale=0.44]{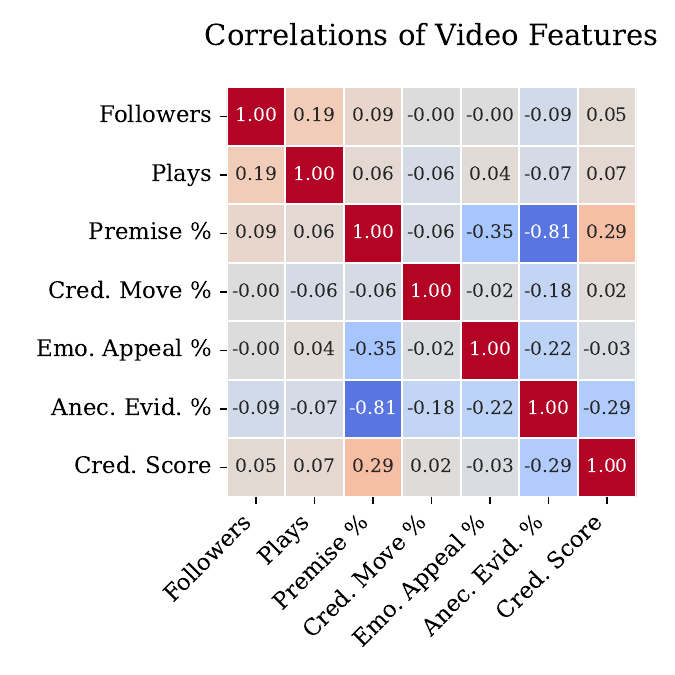}
  \small \caption{\small Is viral content more credible? We plot the correlation matrix of video features, showing that engagement metrics are decoupled from credibility (near-0 correlation).}
  \label{fig:tiktok_heatmap}
\end{figure}

Having established that our trust score reliably reflects content veracity, we apply \textbf{TAIGR} to all $1{,}430$ influencer videos to analyze how influencer popularity, rhetorical strategy, and credibility interact. Throughout this section, we use \emph{credibility} as a shorthand to denote the \textit{trustworthiness} score assigned by our model.

\noindent \textbf{Popularity $\neq$ Credibility.}
Figure~\ref{fig:tiktok_heatmap} reports correlations between video features and credibility. We observe a \textbf{near-zero correlation between our credibility score and engagement metrics}$-$\# Followers ($r=0.05$), \# Plays ($r=0.07$), indicating that neither account popularity nor video virality is predictive of medical claim validity. In other words, evidence-backed health advice is no more likely to be promoted or consumed than low-credibility content, suggesting that platform dynamics are largely agnostic to veracity.

\noindent \textbf{Anecdotes Reduce Credibility.}
Rhetorical structure is a significant predictor of a video's trustworthiness. We find that Credibility Score correlates positively with the use of Premises ($r=0.29$) and negatively with Anecdotal Evidence ($-0.29$). This suggests that videos grounded in logical claims are inherently more credible than those relying on personal testimonials. While these techniques are often used in opposition ($r=-0.81$), their individual impact on veracity is also opposite.

\noindent \textbf{Influencers vs.\ Experts.}
To contextualize our findings, we compare the influencer videos against the ``gold standard'' Cleveland Clinic (CC) videos. Figure \ref{fig:tiktok_cc_comparison} illustrates the rhetorical divide. The expert content (CC) is dominated by premises (Logos), resulting in a high-credibility distribution. Conversely, the influencer distribution is shifted heavily toward anecdotal evidence (Ethos), confirming that reliance on personal testimonials over structural arguments drives the credibility gap between the two, per our model. Appendix~\ref{app:health-topics} shows further analysis by specific health topics.

\vspace{-0.1cm}
\section{Conclusion}
\vspace{-0.1cm}
We introduce \textbf{TAIGR}, a structured framework for validating influencer content by modeling the \textit{takeaway}, its supporting arguments, and external evidence grounding. TAIGR outperforms strong baselines, particularly on videos with implicit recommendations, while enabling interpretable analysis.

% We introduced \textbf{TAIGR}, a structured method for validating influencer discourse on health by modeling the \textit{takeaway}, the argumentation supporting it, and its grounding in external evidence. \textbf{TAIGR} outperforms strong baselines, especially for videos with implicit recommendations, and enables interpretable analysis of influencer content.

\section*{Ethics Statement}
This work studies health information in influencer discourse using transcripts derived from publicly available TikTok videos. We focus on public accounts and collect from influencers with an average $\sim$345{,}000 followers per influencer, aiming to analyze high-reach public-facing content rather than private individuals. We do not access private accounts, bypass access controls, or collect deleted content.

\paragraph{Privacy and identifiability.}
To minimize privacy risks associated with multimedia data, our modeling and analysis operate only on textual transcripts (generated via Whisper) and do not use or release raw video, audio, images, or biometric signals (e.g., faces or voices). We avoid presenting identifying details about individual creators and report findings in aggregate.

\paragraph{Data use, redistribution, and platform policies.}
Platform terms may restrict automated collection and redistribution of content. We therefore do not publicly release raw TikTok content or direct copies of transcripts. To support reproducibility, we will provide access to derived textual data to researchers upon request and agreement not to redistribute.

\paragraph{Potential harms and misuse.}
Studying misinformation may inadvertently amplify harmful claims or enable adversarial use. We mitigate this by (i) focusing on analysis and validation rather than promoting specific creators or claims, (ii) avoiding the release of content that would facilitate re-sharing of misinformation, and (iii) grounding validation in reputable biomedical literature (PubMed). We believe the public-health benefit of understanding and detecting persuasive health misinformation outweighs the residual risks under these safeguards.

\paragraph{Medical disclaimer.}
This work is intended for research; it does not provide medical advice, and model outputs should not be used as clinical guidance.

\section*{Limitations}

\textbf{Retrieved evidence.} Our framework retrieves external evidence exclusively from PubMed. This design choice prioritizes source reliability and consistency in the health domain, but it also limits coverage. Restricting retrieval to a single curated source may exclude relevant evidence available in other scientific repositories or authoritative guidelines. Expanding retrieval to the open web introduces additional challenges, including variability in source trustworthiness and the need to model it, as well as dependence on evidence curated by human fact-checkers~\cite{chen-etal-2024-complex}. Addressing these trade-offs remains an important direction for future work.

\textbf{Text-only modality.} Our analysis operates solely on textual transcriptions of influencer videos. While this enables understanding influencer content to some degree, it ignores potentially informative cues from other modalities, such as visual demonstrations, on-screen text, tone, and gestures. Prior work~\cite{shang_multitec_2025} has shown that multimodal signals can play an important role in interpreting influencer discourse, especially in health misinformation. Incorporating multimodal reasoning remains an important direction for extending the framework.

\textbf{Modular design.} Our framework is intentionally modular, enabling fine-grained inspection of intermediate representations such as takeaways and argumentation structures. While this design introduces the possibility of error propagation across stages, it also allows errors to be explicitly identified, analyzed, and addressed. We address this issue through human evaluation of intermediate components of our framework.

\textbf{Dataset limitation.} There is a lack of publicly available, large-scale datasets specifically designed to analyze influencer discourse and its pragmatic structure. As a result, we construct a proxy dataset aligned with \texttt{ScienceFeedback}, which provides expert-annotated assessments of scientific validity. While this enables systematic evaluation, it may not fully capture the diversity of influencer styles, platforms or non-health domains encountered in real-world settings.

\textbf{Modeling Assumption.} Motivated by the short-form nature of most influencer videos, we assume that each video consists of only one dominant takeaway. However, longer or more complex content may convey multiple recommendations or nuanced messages. Extending the framework to jointly model multiple takeaways within a single video is an important direction for future work.
\section{Statement of AI Assistant Usage}
\label{sec:ai-assistant}
The development of the codebase was partially supported by an AI assistant for debugging, and the manuscript was refined with an AI assistant.

% \section*{Acknowledgments}

\bibliography{custom}

\appendix
\section{Extended Results}
\label{app:extended-results}

\subsection{Impact of Takeaway Type}
\label{app:extended-results-takeaway-type}
Table~\ref{tab:loki_failure_analysis} shows the comparison between LOKI, a claim-centric verifier, against our structured approach towards content validation of influencer discourse. In the implicit cases, we observe that LOKI typically fails to identify the \textit{takeaway}; in this case, fails to associate COVID-19 with ivermectin, and this in turn fails to retrieve relevant evidence, resulting in incorrect classification.
\begin{table}[t!]
\centering
\scriptsize
\resizebox{\columnwidth}{!}{%
\begin{tabular}{p{0.5\columnwidth}|p{0.5\columnwidth}}
\toprule
\textbf{Baseline (LOKI)} & \textbf{TAIGR} \\
% \textit{Focus: Explicit details \& Surface entities} & \textit{Focus: includes implicit takeaway} \\
\midrule
\textbf{Decomposed Claim:} \par The nursing friends and doctor recommended ivermectin. \par \medskip
\textbf{Generated Retrieval Queries:}
% \vspace{-0.5em}
\begin{itemize}[leftmargin=1em, nosep]
    \item What are the clinical indications for ivermectin use according to peer-reviewed guidelines?
    \item Is there evidence that supports recommendations by nursing staff?
\end{itemize} 
& \textbf{Extracted Claim:} \par \textcolor{bluehighlight}{Ivermectin} is recommended by professionals for \textcolor{bluehighlight}{treating COVID symptoms}. \par\medskip
\textbf{Generated Retrieval Queries:}
% \vspace{-0.5em}
\begin{itemize}[leftmargin=1em, nosep]
    \item Clinical trials of \textcolor{bluehighlight}{ivermectin} prescribed by doctors for \textcolor{bluehighlight}{COVID-19 symptoms}.
    \item Adverse effects reported from \textcolor{bluehighlight}{ivermectin} use in \textcolor{bluehighlight}{COVID-19 patients}.
\end{itemize} 
% \\ 
% \midrule
% \textbf{Decomposed Claim:} The person was diagnosed with COVID at CRMC urgent care. \par\medskip
% \textbf{Generated Retrieval Questions:} 
% \vspace{-0.5em}
% \begin{itemize}[leftmargin=1em, nosep]
%     \item Is CRMC urgent care a recognized facility for COVID-19 testing and diagnosis?
% \end{itemize}
% & \textbf{Extracted Claim:} \par
% Taking ivermectin may relieve COVID-19 symptoms.
\\
\bottomrule
\end{tabular}%
}
\caption{\small Comparison of LOKI (baseline) against \textbf{TAIGR} for the \textit{transcript-1} (Table~\ref{tab:takeaway_sample_with_type}). None of the LOKI-decomposed claims account for the implication of \textit{treating COVID-19} with ivermectin. Consequently, it fails to correctly classify the transcript when the takeaway type is \textit{implicit}.}
% evidence retrieval focus. LOKI (left) decomposes claims based on explicit entities, failing to include the \textit{implicitly} mentioned takeaway. Our method (right) correctly identifies the relevant claims, leading to relevant evidence verification.}
\label{tab:loki_failure_analysis}
\end{table}

\subsection{Human Validation Setup}
\label{app:extended-results-human-validation}
We conduct a human validation study on a randomly sampled set of $100$ videos to assess the accuracy of LLM predictions in: (i) identifying the correct \textit{takeaway}, (ii) determining whether the takeaway is expressed \textit{explicitly} or \textit{implicitly}, and (iii) assigning rhetorical function labels to individual statements in the transcript.

Annotations were collected via a Qualtrics survey. Annotators were first provided with a definition of a \textit{takeaway}, followed by the transcript of a video and the LLM-predicted takeaway. They were asked: \emph{``Is this a valid takeaway from the video?''} (Yes/No). When the takeaway was judged valid, annotators were subsequently asked: \emph{``Is the predicted takeaway type (explicit or implicit) consistent with the video?''} (Yes/No).

Independently of takeaway evaluation, annotators were also shown a random subset of $3-5$ statements from each transcript, together with their predicted rhetorical function labels, and were asked to assess whether the labels were appropriate.

Across the annotated sample, the LLM correctly identified the takeaway in $75\%$ of cases, classified explicit versus implicit takeaways with $79\%$ accuracy, and assigned rhetorical function labels with $72\%$ accuracy. 
% These results indicate that, despite the noisy and conversational nature of influencer discourse, LLMs can reliably recover the high-level semantic intent and rhetorical structure required by our framework.

In this case, human annotators were graduate-STEM students under the age of 40, who were not the authors of the paper.
% \begin{figure}[b!]
% \includegraphics[scale=0.55]{Figures/framework_component_human_validation.pdf}
%   \caption{\small Human evaluation: reports accuracy for takeaway identification, implicit vs. explicit classification, and rhetorical function labeling on a randomly sampled $100$ transcripts set.}
%   \label{fig:human-eval-precision}
% \end{figure}

\subsection{Analysis By Health Topic}
\label{app:health-topics}
\begin{figure}[t!]
\includegraphics[width=0.48\textwidth]{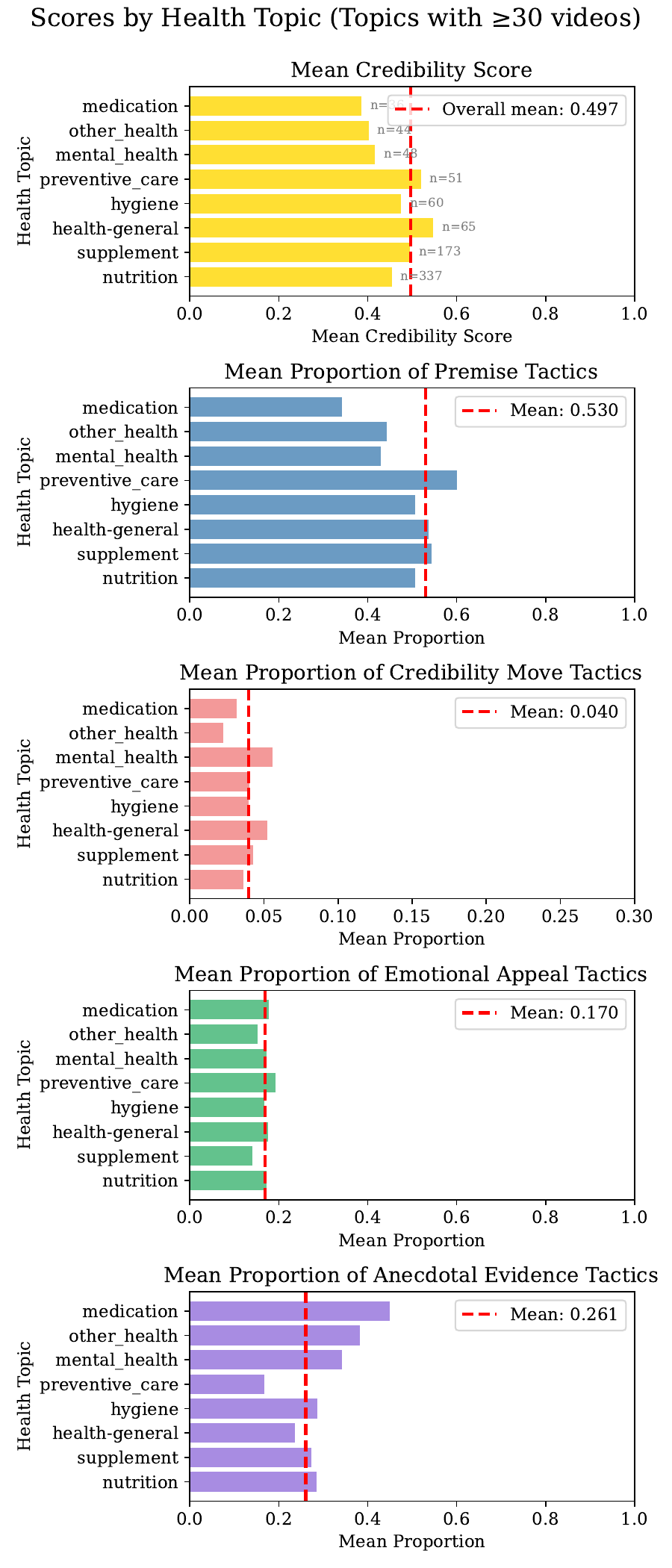}
  \small \caption{\small We show the top Health Topics and their various scores for the full set of $1{,}430$ videos.}
  \label{fig:scores-by-health-topic}
\end{figure}

In Figure~\ref{fig:scores-by-health-topic}, we show all of our framework's average scores for the full set of videos, separated by the health topic. The health topic is determined by the original hashtag(s) used to obtain each video.

We see that for credibility score itself, topics like General Health and Preventative Care are seen as more credible. We hypothesize that this is because these are topics that real medical professionals are willing to provide advice for over social media videos. We see that topics of Medication scores much lower, indicating that many influencers may be making dubious drug recommendations. We see this especially in the Anecdotal Evidence plot, where Medication has a very high proportion of anecdotes. 

\section{Hyperparameters/Implementation Details}
\label{app:sec-methodology-related}
Section~\ref{app:subsec-hyperparameters-implmentation-details} provides the hyperparameters and implementation details associated with each stage in our pipeline.

\subsection{Dataset Curation.}
\label{app:subsec-hyperparameters-implementation-details-dataset}
We describe the detailed procedure used to construct the TikTok health influencer dataset summarized in Section~\ref{sec:data-sec}. We adopt a hashtag-based data collection strategy, where the hashtags are based on the \texttt{ScienceFeedback} website.

Specifically, for each \texttt{ScienceFeedback} article, we provide an LLM with the article text and the main \textit{claim} under evaluation, and prompt it to generate up to five hashtags that are likely to be used by social media users when discussing the claim. These hashtags are intended to capture colloquial terminology commonly employed in influencer discourse.

Using the resulting hashtag set, we collect TikTok videos posted by creators with at least $1{,}000$ followers, ensuring that the dataset focuses on influencer-generated content with non-trivial audience reach. Across all hashtags, we aggregate the retrieved videos and remove duplicates.

We obtain text transcriptions for each video using OpenAI’s Whisper~\citep{10.5555/3618408.3619590}  API. To ensure topical relevance, we prompt an LLM to filter out videos that: (a) do not discuss health-related content, and (b) are not in English language. After this filtering step, the final dataset consists of $1{,}430$ health-related TikTok videos whose transcriptions are in English. Note that we use TikTok's Research API for scraping the videos related to each hashtag.

The resulting corpus exhibits a diverse range of influencer popularity levels and communication styles. Notably, a substantial fraction of videos convey health advice implicitly rather than through explicit recommendations, motivating the takeaway-centric modeling approach proposed in this work.

Details regarding the construction of the annotated evaluation subset, including quality control, are provided in Appendix~\ref{app:subsec-constructing-annotated-dataset}.
% By providing each page from \texttt{ScienceFeedback} along with the main claim for which they have predictions for, we prompt an LLM to come up with most probable set of hashtags (up to $5$) that people in social media typically use when sharing such content online. We use these hashtags to search for influencer content on TikTok, with at least a $1{,}000$ followers. From the extracted videos for all hashtags, we use OpenAI's Whisper API to obtain text transcriptions of each video. We then prompt an LLM to identify if the videos are discussing health-related content or not. Based on filtering for health content, we obtain the set of $1{,}430$ videos. Details regarding the construction of annotated dataset can be found in Appendix~\ref{app:subsec-constructing-annotated-dataset}.

\subsection{TAIGR}
\label{app:subsec-hyperparameters-implmentation-details}
% Here, we provide all the hyperparameters and implementation details needed to reproduce the results from our pipeline.

\paragraph{\textit{Argumentation Structure}.} While constructing the argumentation graph, we treat the \textit{edge weights} between different node pairs differently, ensuring the overall quality of resulting influencer reasoning chains for the \textit{takeaway}. Specifically, we accept a candidate edge between a claim-node and the \textit{takeaway}-node if its edge weight is greater than a threshold of $0.5$, ensuring only strong directly supporting nodes to be associated with the \textit{takeaway} while ignoring noise. This threshold is lowered to $0.25$ when dealing with an edge between a pair of claim-nodes, accounting for those potentially implicit claims, which the influencer indirectly employs while arguing for the takeaway. For instance, a claim node: ``this remedy eliminates worms'' should be connected by an indirectly supporting claim node: ``worms are harmful and are worth eliminating''. Finally, we set a high threshold of $0.5$ for a pair of statement nodes to ensure a high-quality reasoning chain while ignoring noise from the video transcriptions. 

\paragraph{\textit{Trust Inference}.} Implementation details and hyperparameters for this stage of our pipeline are provided below.

\textbf{PubMed data indexing.} We index all PubMed articles from the last decade--$2014$ to $2025$, for augmenting our argumentation graph with trust-worthy scientific biomedical evidence. In total, our index consisted of~$\approx 14M$ articles collected from PubMed annual baseline ($2025$)~\footnote{\scriptsize \url{https://ftp.ncbi.nlm.nih.gov/pubmed/baseline/}}, and filtered based on: (a) \textit{language}--restricted our index to English-only language; (b) \textit{date}--restricted our index to articles from the last decade, ensuring it reflects latest medical guidelines and research. 

\textbf{Retrieval.} For data indexing and retrieval, we utilized the codebase from \texttt{RAGatouille}~\footnote{\scriptsize \url{https://github.com/AnswerDotAI/RAGatouille}}, which internally uses \texttt{ColBERTv2}~\cite{santhanam-etal-2022-colbertv2} for data retrieval. After retrieval, we perform re-ranking using~\texttt{monot5-base-med-msmarco}, where we employ a threshold of $0.3$ (derived empirically by considering the rejection rate and relevance of the retrieved articles).

% \textbf{Factor-graph based inference.} We represent each node $v \in V$ in the augmented argumentation graph as a random variable $Y_v$ with $11$ states $\{0, 0.1, 0.2, ..., 1.0\}$ with higher values indicating greater trustworthiness. We define unary and binary potentials to enforce global consistency in the graph structure and enable trust propagation from high-trust (PubMed) nodes to the rest of the nodes in the graph, including the \textit{takeaway} node. Through unary potentials, we enforce PubMed evidence nodes to have a high-trust score $Y_v = 1$ by defining a hard constraint. Similarly, for the rhetorical node types \{Anecdotal Evidence, Credibility Move, Emotional Appeal\}, we enforce a low-trust score $Y_v = 0.2$. 
\textbf{Factor-graph based inference.} We represent each node $v \in V$ in the augmented argumentation graph as a random variable $Y_v$ with $11$ states $\{0, 0.1, 0.2, ..., 1.0\}$ with higher values indicating greater trustworthiness. We define unary and binary potentials to enforce global consistency in the graph structure and enable trust propagation from high-trust (PubMed) nodes to the rest of the nodes in the graph, including the \textit{takeaway} node. Through unary potentials, we enforce PubMed evidence nodes to have a high-trust score $Y_v = 1$ by defining a hard constraint. In contrast, for the rhetorical node types \{Anecdotal Evidence, Credibility Move, Emotional Appeal\}, we assign a soft low-trust prior via potentials that peak at $Y_v = 0.2$.

To estimate the influence of support and attack potentials (binary potentials), we estimate $\beta, \gamma$ variables using a small sample. Specifically, we randomly sample $20$ datapoints, which follow the same distribution as the labeled dataset ($70\%$ from \textit{incorrect} class, $15\%$ from \textit{partially correct} class, $15\%$ from \textit{correct} class); and use a grid-search to estimate the best-performing values for both variables $\beta, \gamma$. We used: $[0.1, 0.2, 0.3, ... 1.0, 2.0, 3.0, ..., 10.0]$ grid and found that $\{\beta=0.3, \gamma=5\}$ achieved the highest F1-macro points.

For the final classification, we consider the assignment on the \textit{takeaway} node. If the score is in the range $[0, 0.5]$, we classify the input transcription as \textit{incorrect}. If it is in the range $[0.5, 0.7]$, we classify the transcription as \textit{partially correct}. Beyond $0.7$, we classify the transcription as \textit{correct}.

\paragraph{\textit{Other Details.}} We use \texttt{GPT-4.1-mini} with \{$\text{temperature}=0.25, \text{top\_p}=0.95$\} for every LLM call in our pipeline.

\section{Experimental Setup}

\subsection{Constructing Annotated Dataset.}
\label{app:subsec-constructing-annotated-dataset}
To evaluate the models on the \textit{content validation} task, we construct a labeled dataset, using expert annotations from \texttt{ScienceFeedback} as a proxy. \texttt{ScienceFeedback} uses medical experts to validate real-world claims. Often times, it includes ``viral'' claims from online social media such as TikTok or YouTube. After validating a claim, medical experts provide a verdict, typically belonging to one of the pre-defined $18$ classes.

We develop a labeled dataset by matching the main recommendation or \textit{takeaway} from an influencer discourse with the annotated claim from \texttt{ScienceFeedback}. First, we embed all the annotated claims and set of all takeaways associated with each influencer transcription using \texttt{all-mpnet-base-v2}\footnote{\scriptsize \url{https://huggingface.co/sentence-transformers/all-mpnet-base-v2}}~\cite{reimers-gurevych-2019-sentence, song2020mpnetmaskedpermutedpretraining}. For each takeaway, we identify the most similar annotated claim based on cosine similarity, which is above a pre-defined threshold of $0.6$. For the pairs that meet the thresholding criteria, we prompt an LLM to validate if the annotated claim entails the \textit{takeaway}. Specifically, we use the prompt detailed in Figure~\ref{app:subsubsec-prompts-constructing-labeled-dataset} for the classification. By doing this, we identified $185$ potentially semantically equivalent \{takeaway, annotated claim\} pairs.

To ensure semantic equivalence and reliability, we perform human annotation over these pairs to discard the ones where the takeaway does not accurately correspond to the annotated claim, resulting in $166$ high-quality pairs. The labels for the annotated claims in \texttt{ScienceFeedback} is generally skewed and is dominated by \textit{Inaccurate/Unsupported} class, and the predefined $18$ classes creates a class imbalance. Therefore, we collapse the $18$ classes into $3$ classes. Specifically, we map: (1) \{\textit{inaccurate, incorrect, flawed reasoning, unsupported, misleading}\} classes from \texttt{Sciencefeedback} to \textit{\textbf{incorrect}} class; (2) \{\textit{lacks context, imprecise, partially correct}\} classes to \textit{\textbf{partially correct}} class; and (3) \{\textit{mostly accurate, mostly correct, accurate, correct}\} to \textit{\textbf{correct}} class.

Consequently, among the $166$ pairs, we identified $138$ pairs belonged to \textit{incorrect} class, $41$ pairs belonged to \textit{partially correct} class, and $16$ pairs belonged to \textit{correct} class. To increase the coverage of \textit{correct} examples, we augment the dataset with $29$ additional video transcripts sourced from \texttt{Cleveland Clinic}, a medically credible authority, which are all labeled as \textit{correct}. Therefore, in total, our evaluation dataset consists of $195$ pairs: $138$ incorrect pairs, $41$ partially correct pairs and $45$ fully correct pairs. The influencer transcription gets the same label as that of the takeaway.

Aside, in this case, we note that the human annotator was a graduate STEM-student, who was not an author of the paper and was under the age of 30.

\subsection{Baselines.} 
\label{app:subsec-baselines}
Given the healthcare domain of influencer transcripts, we avoid setups that rely on unconstrained end-to-end prompting without external evidence. Such approaches are known to hallucinate or rely on unverifiable reasoning~\cite{lin-etal-2022-truthfulqa}. Instead, all baselines retrieve biomedical evidence from PubMed, ensuring that validation decisions are grounded in scientific literature.

\textbf{Claim-centric RAG.}
This baseline decomposes a transcript into factual claims using an LLM, retrieves the top-$30$ PubMed articles for each claim, and classifies the retrieved evidence as supporting or opposing. The LLM then predicts the overall veracity of the transcript based on the aggregated evidence. This setup mirrors common claim-centric verification pipelines but does not explicitly model the central recommendation conveyed by the influencer.

\textbf{RAG-w-Takeaway.}
This baseline introduces limited structure by first extracting the video’s \textit{takeaway} and retrieving the top-$30$ PubMed articles relevant to it. The retrieved evidence is classified as supporting or opposing, and the LLM is prompted to predict the veracity of the \textit{takeaway} given this evidence. This baseline isolates the effect of takeaway-aware retrieval while omitting explicit modeling of claims and their interactions.

% \textbf{LOKI.}
% We adapt LOKI~\cite{li-etal-2025-loki}, a claim-centric fact verification system, to our dataset. LOKI decomposes transcripts into individual claims, filters uncheckworthy claims, expands checkworthy claims into optimized retrieval queries to improve recall, retrieves evidence from PubMed (in place of its original Google Search API), and verifies each claim independently. Following~\cite{li-etal-2025-loki}, we use their reported threshold for the \textit{incorrect} class and tune the threshold for the \textit{partially correct} class (Appendix~\ref{app:subsec-baselines}).

\noindent \textbf{LOKI.} We adapt~\cite{li-etal-2025-loki} (open-sourced) as a claim-centric baseline, which aggregates the evidence from verifying each claim in order to verify the entire document. Our implementation is based on their official repository\footnote{\scriptsize \url{https://github.com/Libr-AI/OpenFactVerification}}. However, when retrieving evidence for a query, rather than using \texttt{SERPER-API}, we use our retrieval pipeline to obtain evidence from PubMed data, ensuring a fair comparison across all methods. Additionally, we note that~\cite{li-etal-2025-loki} originally used it for binary classification of a claim or a document. Specifically,~\cite{li-etal-2025-loki} used a threshold of $0.8$, below which any claim was classified \textit{incorrect}. We adapt this thresholding to adapt for the 3-class classification problem. By fixing the incorrect class threshold to $0.8$, we empirically obtain a threshold for the \textit{partially correct} class. Performing a grid-search over $[0.8, 0.81, 0.82, ..., 0.99]$, we found that a threshold of $0.95$ achieves the highest F1-macro points, making it the threshold for \textit{partially correct} class.
\section{Statistical Significance Test}
\label{app:stat-significance}
\paragraph{Impact of takeaway type.} We perform a paired bootstrap test ($10,000$ iterations) to compare our method against LOKI (the second-best model). For the implicit takeaway type, this resulted in a p-value of $0.022$, indicating a statistically significant improvement. 

However, for the explicit takeaway type, the test yielded a p-value of $0.136$, indicating that the performance difference is not statistically significant. This indicates that while both models perform comparably on the classification task, our method offers a distinct advantage in scenarios requiring deeper inference regarding influencer's argumentation structure.
\section{Prompt Templates}

\subsection{Our Methodology}
\label{app:subsec-prompts-pipeline}
In this section, we provide all the prompt templates used in each stage of our pipeline: (1) Section~\ref{app:subsubsec-prompts-takeaway} shows the prompt templates used in \textit{takeaway} identification stage; (2) Section~\ref{app:subsubsec-prompts-argumentation-graph} shows the prompts used to model \textbf{how} the influencer argues for the takeaway; (3) Section~\ref{app:subsubsec-prompts-trust-inference} shows the prompts used during the trust inference stage.

\subsubsection{\textit{Takeaway}: What is the influencer trying to communicate?}
\label{app:subsubsec-prompts-takeaway}
Figure~\ref{prmpt-fig:takeaway-extraction},~\ref{prmpt-fig:takeaway-classification} shows the prompts used for takeaway extraction and classification, respectively.

\begin{promptbox}[prmpt-fig:takeaway-extraction]{\textit{Takeaway} extraction.}### Task:
You are a text analyzer. From the given transcript, extract exactly **ONE key takeaway** that captures the most important recommendation, suggestion, protocol, or advice for the audience to remember.

### Definitions:
- *Key takeaway:* a single, complete piece of advice or instruction (explicit or implicit).
- *Explicit advice:* direct instruction (e.g., "Exercise 30 minutes daily", "Take vitamin D").
- *Implicit advice:* indirect or suggestive phrasing that still nudges the audience toward action (e.g., "This worked for me", "Hope this helps everyone", "Studies suggest coffee reduces risk of X"). If it pushes the audience to act, treat it as a takeaway.

### Task Rules:
1. Extract exactly ONE *complete* key takeaway.
   - If multiple exist, choose the one most strongly directed at the audience.
   - If equally strong, choose the first one mentioned.
2. Ensure *contextual completeness*.
   - The **takeaway must be understandable in isolation** (no missing references like "this" or "that").
   - Rewrite slightly if needed for clarity, but keep meaning intact. 
   - It should be complete and self-contained without any co-references.
3. Provide a justification.
   - Explain why this is the most important takeaway, referencing the transcript context.
4. Perform grounding.
   - Return the **exact substring(s)** from the transcript that contains the takeaway.
5. Fallback mechanism.
   - If no valid explicit or implicit takeaway is present, output the fallback JSON.

### Strict JSON Response Format:
```json
{{
  "takeaway": {{
    "text": "<insert-complete-takeaway-here>",
    "justification": "<why this takeaway is important>",
    "grounding": {{
      "text": "<exact-text-span(s)>"
    }}
  }}
}}
```
**If you cannot find a valid takeaway, respond with the following *fallback JSON structure*:**
```json
{{
  "takeaway": "NO_TAKEAWAY_FOUND"
}}
```

### Input Transcript:
{transcript}
\end{promptbox}

\begin{promptbox}[prmpt-fig:takeaway-classification]{\textit{Takeaway} classification.}### Task:
You are a text analyzer. You are also provided with a `takeaway`, which captures the most important recommendation, suggestion, protocol, or advice for the audience to remember from the provided `transcript`. Your task is to classify if the `takeaway` is "explicit" or "implicit" with respect to the `transcript`.

### Definitions:
- *Key takeaway:* a single, complete piece of advice or instruction (explicit or implicit).
- *Explicit advice:* direct instruction (e.g., "Exercise 30 minutes daily", "Take vitamin D").
- *Implicit advice:* indirect or suggestive phrasing that still nudges the audience toward action (e.g., "This worked for me", "Hope this helps everyone", "Studies suggest coffee reduces risk of X"). If it pushes the audience to act, treat it as a takeaway.

### Task Rules:
1. Read the `transcript` and `takeaway` carefully. If the `takeaway` is due to **explicit advice**, classify it as "explicit". If the `takeaway` is due to **implicit advice**, classify it as "implicit".
2. Provide a justification.
   - Explain why this classification is correct by providing an explanation that is grounded to the `transcript`.

### Strict JSON Response Format:
```json
{{
    "label": <"explicit" or "implicit">,
    "justification": <why this takeaway is "explicit" or "implicit" by providing evidence from the transcript>,
}}
```

### Input:
- **`Transcript`**: {transcript}

- **`Takeaway`**: {takeaway}
\end{promptbox}

\subsubsection{\textit{Argumentation Structure}: How does the influencer support the takeaway?}
\label{app:subsubsec-prompts-argumentation-graph}
Figures~\ref{prmpt-fig:arg-graph-preprocess-standalone-statements},~\ref{prmpt-fig:arg-graph-preprocess-classification},~\ref{prmpt-fig:arg-graph-preprocess-claim-extraction} shows the prompts used in the \textit{preprocessing} phase. Figure~\ref{prmpt-fig:arg-graph-construction} shows the prompt used in constructing the argumentation graph: identifying the potential support relationship between two nodes (between a claim-claim node, claim-takeaway node, and statement-statement node). A statement node can be one of the following \{\textit{Premise, Anecdotal Evidence, Credibility Move, Emotional Appeal}\}.

\begin{promptbox}[prmpt-fig:arg-graph-preprocess-standalone-statements]{\textit{Preprocessing}: Obtaining standalone statements from the video transcriptions.}### Task:
You are an expert in text analysis. Your task is to rewrite the provided `text` as a list of **completely standalone sentences**, where each sentence can be understood on its own while preserving the original tone and point of view.

### Strict Requirements:
1. **Sentence Splitting:** Split the `text` into individual sentences.
2. **Coreference Resolution:** 
  - **Preserve Point of View:** Maintain the original first-person perspective ("I," "my," "me"). Do **not** replace these with "the speaker."
  - **Resolve Ambiguity:** Replace pronouns (e.g., “he,” “it,” “they”) or vague references with the correct noun phrase from the `text`, so that each sentence is **fully self-contained**.
3. **Grounding:** Do not add, remove, or invent any information. For each output sentence, strictly ensure that it is **fully grounded** in the `text`.
4. **Inclusion:** Every sentence in the original `text` must appear in the output.

### Output Format (Strict JSON):
```json
{{
  "all_standalone_sentences": [
    "<completely standalone sentence strictly grounded in the provided `text`>",
    ... # strictly include **all** sentences from the `text`, no information should be omitted
    ]
}}
```

### Example:
**`Input Text`:** Okay, I know this is gonna sound wild — but your own urine might be one of the most powerful natural boosters out there. Think about it: it's made by you, packed with your own minerals, hormones, and amino acids — literally personalized medicine. I've been using aged urine topically for years, just rubbing a little on my skin before going out in the sun. I swear, the warmth activates it — my muscles feel fuller, my energy spikes, and my skin gets this crazy healthy glow. And yeah, sometimes I'll even take a sip of my morning urine. Sounds gross, but the energy and clarity I get after? Unreal. I'm almost 60, don't use sunscreen or steroids, and I still feel strong and alive. It's not for everyone, but for me, it's been a total game-changer. Sometimes the weirdest things nature gives us are the ones that actually work.

**`Output JSON`:**
```json
{{
  "all_standalone_sentences": [
    "Okay, I know this is gonna sound wild — but your own urine might be one of the most powerful natural boosters out there.",
    "Your urine is made by you and is packed with your own minerals, hormones, and amino acids — literally personalized medicine.",
    "I have been using aged urine topically for years by rubbing a little of the aged urine on my skin before going out in the sun.",
    "I swear that the warmth of the sun activates the aged urine.",
    "When the warmth activates the aged urine, my muscles feel fuller, my energy spikes, and my skin gets this crazy healthy glow.",
    "Sometimes I even take a sip of my morning urine.",
    "Drinking my morning urine sounds gross, but the energy and clarity I get after drinking it are unreal.",
    "I am almost 60 years old, do not use sunscreen or steroids, and I still feel strong and alive.",
    "Using your own urine is not for everyone, but for me, using it has been a total game-changer.",
    "Sometimes the weirdest things that nature gives us are the ones that actually work."
  ]
}}
```
**From the example output JSON, it is important to note that every sentence from the input text is included, every sentence is standalone, preserving sentence-level self-containment, and original tone and point of view.**

### Input:
`text`: {transcript}
\end{promptbox}

\begin{promptboxwide}[prmpt-fig:arg-graph-preprocess-classification]{\textit{Preprocessing}: Classifying each statement based on its rhetorical functionality.}### Task:
You are an expert in argumentation theory and discourse analysis. Your goal is to classify each provided `Statement` into **exactly one** of the following categories, based on its rhetorical function and style of persuasion. Understand the provided `Transcript` to help you classify the statements.

---

### Classification Categories:

#### 1. "Premise" (Logos)
- **Definition:** A logical or factual statement that provides *reason-based* support for a claim.
- **Key Features:**
  - Uses reasoning, causation, or factual explanation ("because", "therefore", "since").
  - Does **not** rely on personal experience, emotion, or authority.
- **Example:** "Because urine contains urea and minerals, it may have antimicrobial properties."
- **Not Premise if:** The statement talks about personal experience, credentials, or emotional outcomes.

---

#### 2. "Anecdotal Evidence"
- **Definition:** A personal story, observation, or example based on individual experience rather than systematic proof.
- **Key Features:**
  - Mentions the speaker's or someone's *own experience* ("I tried", "my friend noticed", "people say").
  - Often used to illustrate or reinforce a point with a story.
- **Example:** "I started drinking urine last year, and my skin has never been clearer."
- **Not Anecdotal if:** The statement generalizes or reasons logically without a personal element.

---

#### 3. "Credibility Move" (Ethos)
- **Definition:** A statement that builds trust, expertise, or authority of the speaker to make the argument more convincing.
- **Key Features:**
  - References credentials, background, authority, or expertise ("As a doctor...", "I've studied this for years", "According to this research...").
  - Sometimes includes name-dropping or citing authoritative sources.
- **Example:** "As a practitioner of Ayurvedic medicine, I can assure you this technique is safe."
- **Not Credibility Move if:** It focuses on data, reasoning, or emotions rather than authority.

---

#### 4. "Emotional Appeal" (Pathos)
- **Definition:** A statement that evokes emotion (hope, fear, pride, shame, love, empowerment) to persuade.
- **Key Features:**
  - Uses emotionally charged language or imagery.
  - Focuses on how the audience *feels* rather than what is *true*.
- **Example:** "It's time to reclaim control of your body — don't let big pharma tell you what's natural."
- **Not Emotional Appeal if:** The argument is calm, factual, or authority-based.

---

#### 5. "None" (None)
- **Definition:** The statement doesn't clearly fit any of the categories above.
- **Examples:**
  - Off-topic comments or factual statements unrelated to persuasion.
  - Ambiguous fragments without clear function.

---

### Additional Guidelines:
1. **Prioritize rhetorical intent** over surface wording — ask: *What is the statement trying to do?*
2. **Avoid defaulting to "Premise"** unless the statement clearly relies on reasoning.
3. **When authority-based — prefer "Credibility Move".**
4. **When personal or experiential — prefer "Anecdotal Evidence".**
5. **When emotionally loaded — prefer "Emotional Appeal".**
6. **When unclear — use "None".**

---

### Strict JSON Output Format:
```json
{{
    "<statement_id>": {{    # `statement_id` should match the `id` of the statement in the given `statements` dict.
        "type": "<classification type>",
        "explanation": "<clear justification referring to its features>"
    }},
    ...
}}
```

---

### Input:
`**Transcript**`: {full_transcript}

---

`**Statements**`: {statements_dict}
\end{promptboxwide}

\begin{promptboxwide}[prmpt-fig:arg-graph-preprocess-claim-extraction]{\textit{Preprocessing}: Claim identification -- accounting for potentially implicit reasoning chains.}### Task:
You are an expert in structured argumentation analysis. Your goal is to identify **ALL the claims** being made in the given `transcript`. You are also provided with individual `statements` from the same `transcript`, each with a unique `id`. 

### Definition of a Claim:
A **claim** is a statement that asserts a position, viewpoint, or conclusion that can be evaluated as true, false, or debatable. Claims are the central points of an argument. There are two types of claims:

- **Explicit Claim:** A statement directly expressed in the text where the speaker clearly asserts a position or conclusion.
- **Implicit Claim:** A statement that suggests, implies, or presupposes a position without directly stating it; the claim must be inferred from context, assumptions, causal reasoning, or the relationship between statements.

### Instructions:
- Read and understand the `transcript` and the `statements` carefully.
- Identify **both explicit and implicit claims**, including claims that describe causes, effects, assumptions, goals, motivations, or implied problems and solutions.
- For each statement in the transcript:
    - Decide if it expresses (explicitly or implicitly) a claim.
    - If it does, extract it and add it to the dictionary of identified claims.
- If multiple statements together express one single claim, combine them. Add the combined statement to the identified dictionary of claims.
- If a statement(s) is not making a claim, skip it.
- Ensure each identified claim is **different** from the others. We do not want the same claims being identified multiple times.
- When inferring **implicit claims**, look for:
    - Assumptions that make explicit claims meaningful (e.g., "worms are harmful" implied by "this remedy eliminates worms"; OR "urine is beneficial for health" implied by “applying urine to the skin improves skin health”).
    - Implied causal or motivational links (e.g., "eliminating worms improves health"; OR "the warmth from sunlight activates beneficial properties in urine" implied by "the warmth activates it - muscles feel fuller and energy spikes").
    - Contextual premises or problem statements that justify a proposed action (e.g., "natural bodily substances can provide health benefits when reused correctly" implied by "your own urine might be one of the most powerful natural boosters out there").
- Each identified claim must be **self-contained** and understandable on its own, without referring back to the transcript.
- For every identified claim, record the following information to remain faithful to the given `transcript`:
    - `claim_id`: The <id> of the claim. Start from 1 and increment for each new claim. Format: `claim_<id>`.
    - `claim_text`: The text of the claim that can be understood on its own without referring to the `transcript`.
    - `claim_type`: The type of the claim ("explicit" or "implicit").
    - `statement_ids`: Include **all** the sentence ids that are part of the identified claim. **Ensure that the `ids` map to the ids of the statements in the `statements` dictionary.** This is important to ensure that the claim is faithful to the transcript.
    - `justification`: Provide a brief justification for why you identified this as a claim and whether it was explicitly stated or implicitly inferred.
    - `grounding_to_statements`: Provide a brief justification for how this claim is grounded to the `statement_ids` in the given `statements` dictionary, including reasoning about inferred or combined meaning if applicable.
- **If you cannot identify even a single claim, then return an empty dictionary.**

### Example showing extracted `claims` from a `transcript`:
**`Transcript`**:
Okay, I know this is gonna sound wild — but your own urine might be one of the most powerful natural boosters out there. Think about it: it's made by you, packed with your own minerals, hormones, and amino acids — literally personalized medicine. I've been using aged urine topically for years, just rubbing a little on my skin before going out in the sun. I swear, the warmth activates it — my muscles feel fuller, my energy spikes, and my skin gets this crazy healthy glow. And yeah, sometimes I'll even take a sip of my morning urine. Sounds gross, but the energy and clarity I get after? Unreal. I'm almost 60, don't use sunscreen or steroids, and I still feel strong and alive. It's not for everyone, but for me, it's been a total game-changer. Sometimes the weirdest things nature gives us are the ones that actually work.

**`Extracted Claims` (showing *only* the `claim_text` and `claim_type`)**:
1. Human urine is a powerful natural booster. (explicit)
2. Urine contains beneficial concentrations of minerals, hormones, and amino acids. (explicit)
3. Urine functions as personalized medicine because it is composed of an individual's own unique minerals, hormones, and amino acids. (implicit)
4. The warmth from sunlight activates the properties of topically-applied urine. (explicit)
5. Applying sun-activated, aged urine to the skin causes muscles to feel fuller, energy to spike, and skin to develop a healthy glow. (explicit)
6. Ingesting one's own morning urine results in increased energy and mental clarity. (implicit)
7. Natural bodily substances can have health benefits when reused correctly. (implicit)

It is important to note that each `claim_text` can be understood on its own without referring to the `transcript`. It does **not** contain any information such as "the speaker" or "I" or "my" or "me" or any other pronouns.

### Strict JSON Output Format:
```json
{{
    "<claim_1>": {{
        "claim_text": "Human urine is a powerful natural booster.", 
        "claim_type": "explicit", 
        "statement_ids": [S1],  
        "justification": "The speaker asserts a specific viewpoint that human urine is a powerful natural booster. It is a debatable assertion that can be evaluated as true or false.", 
        "grounding_to_statements": "S1 consists of the statement 'Your own urine might be one of the most powerful natural boosters out there.' which is a clear assertion that human urine is a powerful natural booster."
    }},
    ...
}}
```
**If you cannot identify even a single claim, then return the following empty dictionary.**
```json
{{}}
```

### Input:
`Transcript`: {transcript}

---

`Statements`: {statements_dict}
\end{promptboxwide}

\begin{promptbox}[prmpt-fig:arg-graph-construction]{Building Argumentation Graph: Identifying relationship between two nodes.}### Task:
You are an expert argument analyst. You are provided with a `Claim to Analyze` and a list of `Available Claims`. For each claim in the list, your goal is to classify it into one of the following categories based on how well it supports the `Claim to Analyze`. The categories are: **direct support**, **weak support**, or **no support**, each of which is defined below. You are also provided with the `Transcript`, which you can use to get additional context about the `Claim to Analyze` and the claims in the list.

If a claim fits more than one category, assign it to the **strongest** one (for example, prefer *direct support* over *weak support* if both could apply; prefer *weak support* over *no support* if both could apply;).

### Definitions:
- **direct support:** Claim A *strongly directly supports* Claim B if A clearly *restates, proves, or provides explicit causal justification* for B, leaving little interpretive ambiguity.
    - Example: `Claim to Analyze` is "Drinking green tea helps with weight loss" and a directly supporting claim is "I lost weight after switching to green tea." or "Green tea boosts metabolism, which helps the body burn fat faster." (explicit causal justification)
- **weak support:** Claim A *weakly supports* Claim B if A is *topically relevant and consistent in stance*, but does not directly assert, prove, or justify B. The support is inferential or anecdotal rather than explicit.
    - This includes claims that highlight the importance, relevance, or consequences of the topic discussed in B. For example, stating that the issue requires attention, action, treatment etc. can increase the plausibility of B without proving it.
    - Example: `Claim to Analyze` is "Drinking green tea helps with weight loss" and an weakly supporting claim is "Lots of people drink it to feel lighter." or "It's part of many healthy diets."
    - Another example: `Claim to Analyze` is "Green tea boosts metabolism and burns fat." and an indirectly supporting claim is "Green tea supports gut health and digestion." (*causal link between digestion and fat-burning is implied, not stated*.)
- **no support:** Claim A *does not support* Claim B if there is *no meaningful logical, causal, or thematic connection* that would make B more believable, plausible, or credible.
    - Example: `Claim to Analyze` is "Drinking green tea helps with weight loss" and a *no support* claim is "I prefer coffee in the morning."

### For Each Selected Claim:
For every claim you select, provide:
-   `"id"` — the claim ID from the available list. It must **strictly** be one of the IDs from the available list.
-   `"category"` — the category of the claim ("direct support", "weak support", or "no support").
-   `"explanation"` — why this claim supports/does not support the `Claim to Analyze` based on the definitions above. The explanation should be complete, concise, and clear.

### Strict JSON Output Format:
If you find supporting claims, return a **valid JSON** with the following format:
```json
{{
    "<claim-id>": {{
        "category": "<category of the claim>",
        "explanation": "<why this claim supports/does not support the `Claim to Analyze`>",
    }}
}}
```

### Input:
`**Transcript**`: {full_transcript}
`**Claim to Analyze**`: {claim_text}
`**Available Claims JSON**`: {available_claims_json}
\end{promptbox}

\subsubsection{\textit{Trust Inference}: Is the influencer's argument accurate?}
\label{app:subsubsec-prompts-trust-inference}
Figures~\ref{prmt-fig:trust-inference-checkworthy-classification},~\ref{prmpt-fig:trust-inference-expanding-search-queries},~\ref{prmpt-fig:trust-inference-classifying-retrieved-evidence} shows the prompts for identification of checkworthy statements, expanding search queries, and classification of retrieved evidence, respectively.

\begin{promptbox}[prmt-fig:trust-inference-checkworthy-classification]{Classifying whether a node is PubMed-checkworthy or not.}### Task:
You are an expert at understanding the provided text. You are provided with a `context` (transcript) and a several texts derived from the `context`. Your task is to classify each provided `text` as either `PubMed-checkworthy` or `not PubMed-checkworthy`.

### Decision Guide:
A text is **PubMed-checkworthy** if:
- It makes a **biomedical or health-related factual claim** that can be evaluated with scientific evidence (e.g., about diseases, treatments, nutrients, physiology, hormones, biological effects, and many more).
- The claim is **generalizable beyond the specific provided context** and could, in principle, be **verified or refuted** using PubMed or biomedical literature.

A text is **not PubMed-checkworthy** if:
- It is **personal, anecdotal, or rhetorical** (e.g., "I love how it makes me feel").
- It is **too vague or context-specific** to be scientifically verified (e.g., "this is aged urine").
- It is **non-biomedical** (e.g., about opinions, beliefs, emotions, aesthetics, or general lifestyle).

### Classification Rules:
1. Analyze the provided `context` to get a sense of the overall context associated with each `text`.
2. For each `text`, decide if it expresses a verifiable biomedical claim.
3. When classifying a `text`,
    - if multiple labels are applicable, prioritize `PubMed-checkworthy` over `not PubMed-checkworthy`.
    - if in doubt, classify as `PubMed-checkworthy`.
4. Output one JSON entry per text, preserving its exact `id`. This `id` must strictly match the `id` of the `text` in the input.
5. Strictly ensure that each `id` provided in the input is present in the output.

### Strict JSON Output Format:
```json
{{
  "<text_id>": {{
    "label": "PubMed-checkworthy" | "not PubMed-checkworthy",
    "explanation": "<brief reason justifying the classification decision>"
  }}
}}
```

### Example:
`Transcript`: "This is my aged urine. Okay, I know this is gonna sound wild — but your own urine might be one of the most powerful natural boosters out there. Think about it: it's made by you, packed with your own minerals, hormones, and amino acids — literally personalized medicine. I've been using aged urine topically for years, just rubbing a little on my skin before going out in the sun. I swear, the warmth activates it — my muscles feel fuller, my energy spikes, and my skin gets this crazy healthy glow. And yeah, sometimes I'll even take a sip of my morning urine. Sounds gross, but the energy and clarity I get after? Unreal. I'm almost 60, don't use sunscreen or steroids, and I still feel strong and alive. It's not for everyone, but for me, it's been a total game-changer. Sometimes the weirdest things nature gives us are the ones that actually work."

`Texts to classify`: {{
    "claim_1": "Aged urine boosts testosterone and improves skin health.",
    "S2": "This is aged urine."
}}

`Output`: {{"claim_1": {{
    "label": "PubMed-checkworthy", 
    "explanation": "This is a biomedical claim about hormonal and skin effects that can be verified using scientific literature."
}},
"S2": {{
    "label": "not PubMed-checkworthy", 
    "explanation": "This is a descriptive statement tied specifically to the provided context, not a biomedical claim that can be checked against PubMed."
}}}}

### Input:
`Transcript`: {transcript}
`Texts to classify`: {nodes_json}
\end{promptbox}

\begin{promptboxwide}[prmpt-fig:trust-inference-expanding-search-queries]{Expanding search queries for checkworthy nodes.}You are an expert at generating short *semantic* search queries for biomedical literature retrieval (for use with embedding-based / semantic search systems). Given a `transcript` (context) and a `claim` already marked PubMed-checkworthy, generate concise, natural-language queries that maximize semantic recall for both **supporting** and **opposing** evidence.

### Output goals:
- Produce up to **5 supporting** and up to **5 opposing** semantic queries.
- Queries must be **natural-language**, standalone, and optimized for embedding-based retrieval (no boolean operators, no Medline field tags).
- Each query should be paired with a one-sentence **justification** explaining what kind of evidence it targets.

### Query construction heuristics (semantic best-practices):
1. **Natural phrasing**: write queries as short sentences or questions a researcher might ask (e.g., "Topical human urine effect on skin hydration in adults?"). Avoid using AND/OR, quotes, field tags, or forced Boolean structure.
2. **Diverse angles**: for both supporting and opposing sets, vary the focus across:
   - Outcome/effect (e.g., testosterone, skin appearance, muscle recovery)
   - Mechanism/biomarkers (e.g., hormones, amino acids, immune markers)
   - Population and application route (topical vs oral; adults vs older adults)
   - Study design/strength (randomized controlled trial, clinical trial, observational study, review/meta-analysis)
   - Safety/adverse events and null findings (no effect, adverse outcomes)
3. **Paraphrase strategy**: include different grammatical forms — short declarative, question form, and mechanism-focused phrase. This increases embedding-space coverage.
4. **Contrast language for opposing queries**: use explicit negation or null-effect language (e.g., "no increase", "no association", "ineffective") and terms like "lack of benefit", "no effect", "adverse" to surface contradictory or null-result papers.
5. **Conciseness**: aim for approximately 5-12 words per query (short enough to get focused embeddings, long enough to include key concepts).
6. **No repetition**: each query must add a new angle or emphasize a different keyword set (in natural language), not just reword the same phrase.
7. **Scope alignment**: all generated queries must stay *strictly within the scope of the provided claim*. Do not introduce new effects, outcomes, mechanisms, or populations not mentioned or implied by the claim. For example, if the claim concerns "Topical aged urine increasing testosterone and improving skin appearance", queries about "reducing brain fog" or unrelated outcomes are invalid.

### Strict Output JSON format:
Return valid JSON:
```json
{{
  "supporting_queries": [
    {{"query_id":"S1","query_text":"<natural-language semantic query>","justification":"<why this finds supporting evidence, and why is the scope aligned with the claim>"}},
    ...
  ],
  "opposing_queries": [
    {{"query_id":"O1","query_text":"<natural-language semantic query>","justification":"<why this finds opposing/null evidence, and why is the scope aligned with the claim>"}},
    ...
  ]
}}
```
If you are unable to generate any supporting queries, return an empty list for the "supporting_queries" key. Similarly, if you are unable to generate any opposing queries, return an empty list for the "opposing_queries" key. If you are unable to generate any queries, return an empty list for both the "supporting_queries" and "opposing_queries" keys.

### Example:
Claim: "Topical aged urine increases testosterone and improves skin appearance."

Supporting queries (examples):
```json
{{"query_id":"S1", "query_text":"Topical urine application and testosterone changes in adults", "justification":"Targets studies measuring hormone changes after topical urine exposure. It is specific to the provided claim. The scope of the query is within the context of the provided claim."}}
{{"query_id":"S2","query_text":"Effects of topical application of human urine on skin hydration and color","justification":"Catches dermatology studies on topical biological secretions and skin outcomes.This is within the context of the provided claim."}}
```
Opposing queries (examples):
```json
{{"query_id":"O1","query_text":"Clinical trials showing no testosterone increase after topical application of human urine","justification":"Finds RCTs or trials reporting no hormonal effects from topical interventions. It is specific to the provided claim. The scope of the query is within the context of the provided claim."}}
{{"query_id":"O2","query_text":"Adverse outcomes or infection risk from applying human bodily fluids such as urine to skin","justification":"Targets safety and adverse-event literature about topical body-fluid use. It is specific to the provided claim. The scope of the query is within the context of the provided claim."}}
```

### Input:
**`Transcript`**: {transcript}
**`Claim`**: {claim}
\end{promptboxwide}

\begin{promptboxwide}[prmpt-fig:trust-inference-classifying-retrieved-evidence]{Classification of retrieved evidence into: \{\textit{strong support, weak support, neutral, weak oppose, strong oppose}\}.}### Task:
You are an expert at natural language inference and scientific stance classification. You are provided with a `Claim` and a list of potentially relevant `evidence articles`. For each evidence in the list, your goal is to classify it into one of the following categories, reflecting how strongly the evidence relates to and aligns with the `Claim`:  
**strong support**, **weak support**, **neutral**, **weak oppose**, or **strong oppose**.
You are also provided with the `Transcript`, which gives broader context about the `Claim` and may help you classify the evidence.

### Definitions:
- **strong support:**
    - Evidence is *directly consistent* with the claim and provides **explicit empirical or causal confirmation** of the same relationship, typically through high-quality experimental, clinical, or meta-analytic findings that show a statistically significant and meaningful effect.
    - Such evidence *affirms* the claim's key variables, directionality, and population with high certainty.
    - *Example:*
        - Claim: "Green tea reduces blood pressure in adults."
        - Evidence: "A large randomized controlled trial found that daily green tea consumption significantly decreased both systolic and diastolic blood pressure in adults."
        - Explanation: **strong support** because of explicit causal agreement from a high-quality study design, affirming the key variables and population.
- **weak support:**
    - Evidence is *broadly consistent* with the claim but provides **indirect, correlational, or partial justification**, rather than direct and complete proof.
    - This includes mechanistic reasoning, findings from different but related populations (e.g., animal models for a human claim), or evidence that confirms only a part of the claim.
    - *Example 1 (Indirect Population):*
        - Claim: "Green tea reduces blood pressure in adults."
        - Evidence: "In a study on hypertensive rats, consumption of green tea extract led to a significant reduction in blood pressure."
        - Explanation: **weak support** as the evidence is from an animal model, offering only indirect, inferential support for the human claim.
    - *Example 2 (Partial Confirmation):*
        - Claim: "Green tea reduces blood pressure in adults."
        - Evidence: "A clinical study observed that green tea consumption was associated with a statistically significant decrease in systolic blood pressure, but not diastolic blood pressure."
        - Explanation: **weak support** because the evidence only partially confirms the claim's outcome (one measure of blood pressure, but not the other).
- **neutral:**
    - Evidence provides **no clear stance** toward the claim's truth, either because it is *topically unrelated* or discusses a *logically irrelevant* aspect of the topic.
    - It neither supports nor contradicts the claim. Any evidence that is not relevant to the claim's central assertion must be classified as neutral.
    - *Example 1 (Topically Related but Irrelevant):*
        - Claim: "Green tea reduces blood pressure in adults."
        - Evidence: "Green tea enhances cognitive alertness."
        - Explanation: **neutral** as the evidence shares the topic but is irrelevant to the claimed relationship with blood pressure.
    - *Example 2 (Topically Unrelated):*
        - Claim: "Green tea reduces blood pressure in adults."
        - Evidence: "Regular cardiovascular exercise is effective for improving heart health."
        - Explanation: **neutral** because the evidence is on a completely different topic and has no bearing on the claim about green tea.
- **weak oppose:**
    - Evidence is *topically relevant* but presents **inconclusive, mixed, or weakly contradictory findings**.
    - It may report a lack of a statistically significant effect from a study that may be underpowered, or present findings with limitations that cast doubt without conclusively refuting the claim.
    - *Example:*
        - Claim: "Green tea reduces blood pressure in adults."
        - Evidence: "A small pilot study with 30 participants found no statistically significant change in blood pressure after four weeks of green tea supplementation."
        - Explanation: **weak oppose** because the findings from a small, potentially underpowered study fail to support the claim but do not have enough statistical power to strongly refute it.
- **strong oppose:**
    - Evidence *directly contradicts* the claim by providing **robust empirical refutation or inverse findings**.
    - This includes conclusive null effects from large, high-powered studies or meta-analyses, or significant findings showing an effect in the opposite direction.
    - *Example:*
        - Claim: "Green tea reduces blood pressure in adults."
        - Evidence: "A meta-analysis of 15 randomized controlled trials, including over 2,000 participants, found no statistically significant effect of green tea consumption on blood pressure."
        - Explanation: **strong oppose** because a large-scale, high-quality evidence synthesis explicitly and conclusively finds a null effect, refuting the claim.

### For Each Evidence:
For every evidence, provide:
- `"id"` — the evidence ID from the available list. It must **strictly** match one of the provided IDs.
- `"category"` — one of the five stance labels (`"strong support"`, `"weak support"`, `"neutral"`, `"weak oppose"`, or `"strong oppose"`).
- `"explanation"` — a concise justification explaining *why* this evidence fits that stance, referencing causal direction, strength of association, and topical relevance. Your explanation should be precise, especially for boundary cases like partial support or indirect evidence from different populations.

### Strict JSON Output Format:
Return valid JSON in the following structure:
```json
{{
    "<evidence-id>": {{
        "category": "<strong support | weak support | neutral | weak oppose | strong oppose>",
        "explanation": "<why this evidence fits this category>"
    }}
}}
```

### Input:
`**Transcript**`: {transcript}
`**Claim**`: {claim_text}
`**Available Evidence**`: {evidence_formatted}
\end{promptboxwide}

\subsection{Experimental Setup}
\label{app:subsec-prompts-experimental-setup}
In this section, we provide all the prompt templates used for constructing a labeled dataset, and developing baselines.

\subsubsection{Constructing labeled dataset}
\label{app:subsubsec-prompts-constructing-labeled-dataset}
\begin{promptbox}[prmpt-fig:constructing-labeled-dataset-llm-equivalence]{Entailment between \{\textit{Takeaway}-annotated claim\} for constructing labeled dataset.}You are an entailment expert. Given `Text-1` and `Text-2`, decide if `Text-2` entails `Text-1`. 

### Key Rule:
- Ignore any recommendations, advice, instructions, or protocols in the texts.
- Focus only on the **core factual content** (claims about properties, causes, or effects).
- `Text-2` entails `Text-1` if everything asserted in `Text-1` is logically implied by `Text-2`.

---

### Examples

#### Example 1:
- Text-1: "Avoid wearing AirPods close to your brain because Bluetooth radiation can cause health issues."
- Text-2: "AirPods emit dangerous levels of radiation that harm the body."
- After removing the advice from Text-1, the core is: "Bluetooth radiation from AirPods can cause health issues."
- Text-2 asserts that AirPods emit harmful radiation. This logically implies the core of Text-1.  
- **Answer:** entails = yes.

---

#### Example 2:
- Text-1: "To support your immune system and help prevent or treat infections during sick season, make a natural remedy called 'Nature's Amoxicillin' by blending onion, ginger, turmeric, garlic, lemon with peel, honey, cayenne, and other potent antiviral and antibacterial ingredients, and consume it as a wellness shot."
- Text-2: "A soup made of garlic, onions, thyme, and lemon can replace the flu shot and cure other illnesses such as the common cold and norovirus."
- Removing the advice from Text-1, the core is: "A mixture of certain natural ingredients (onion, ginger, turmeric, garlic, lemon, etc.) can help prevent or treat infections during sick season."
- Text-2 asserts that a mixture of similar ingredients (garlic, onions, thyme, lemon) can cure illnesses like the common cold.  
- Both describe natural mixtures of overlapping ingredients with the shared claim that these mixtures prevent or cure infections.  
- **Answer:** entails = yes.

---

#### Example 3:
- Text-1: "Taking one tablespoon daily of the prepared mixture containing ginger, red onion, garlic, turmeric, black pepper, lemon or lime, and honey will boost your immune system by 200%, reduce body inflammation, and protect you from a long list of diseases."
- Text-2: "A soup made of garlic, onions, thyme, and lemon can replace the flu shot and cure other illnesses such as the common cold and norovirus."
- Removing the advice from Text-1, the core is: "A natural mixture with garlic, onion, ginger, turmeric, etc., protects against many diseases and strengthens the immune system."
- Text-2 asserts that a natural soup with overlapping ingredients (garlic, onion, lemon) cures illnesses and can replace the flu shot.  
- Both describe natural mixtures with protective health effects, so Text-2 logically implies the general claim in Text-1.  
- **Answer:** entails = yes.

---

### Output Format (strict JSON):
```json
{{
  "entails": "<yes or no>",
  "justification": "<short explanation>",
  "evidence": "<if entails=yes, quote the supporting part of Text-2; else leave empty>"
}}
```

### Input:
`Text-1`: {text_1}
`Text-2`: {text_2}
\end{promptbox}

% \subsubsection{Baselines}
% \label{app:subsubsec-prompts-baselines}

% \label{sec:framework-related}

\end{document}